\documentclass[11pt, a4paper, logo, copyright]{eai}
\usepackage{shlab}

\usepackage[authoryear, sort&compress, round]{natbib}
\usepackage[]{mdframed}

\usepackage{dblfloatfix}

\usepackage{amsmath,amsfonts,bm}
\usepackage{multirow}
\usepackage{subcaption}
\usepackage{wrapfig}

\def\eqref#1{equation~\ref{#1}}

\def\1{\bm{1}}

\DeclareMathAlphabet{\mathsfit}{\encodingdefault}{\sfdefault}{m}{sl}
\SetMathAlphabet{\mathsfit}{bold}{\encodingdefault}{\sfdefault}{bx}{n}

\usepackage{listings}
\usepackage{hyperref}
\usepackage{url}
\usepackage{graphicx}
\usepackage{amsmath} %
\usepackage{mathrsfs} %
\usepackage{etoolbox}
\usepackage{cleveref}
\usepackage{tcolorbox}
\usepackage{colortbl}
\usepackage{booktabs}       %
\usepackage{amsfonts}       %
\usepackage{nicefrac}       %
\usepackage{microtype}      %
\usepackage{caption}
\usepackage{subcaption}
\usepackage{algorithm}
\usepackage{multirow}
\usepackage{lipsum}
\usepackage{algpseudocode}  % 

\usepackage{booktabs} %
\usepackage{enumitem}%
\setlist[itemize]{noitemsep, topsep=0pt}
\usepackage{enumitem,kantlipsum}

\newlength\savewidth
\newcommand{\tablestyle}[2]{\setlength{\tabcolsep}{#1}\renewcommand{\arraystretch}{#2}\centering\footnotesize}
\definecolor{baselinecolor}{HTML}{d6eaf8}
\newcommand{\baseline}[1]{\cellcolor{baselinecolor}{#1}}
\definecolor{mygray}{gray}{0.4}

\AtBeginEnvironment{tcolorbox}{\tiny}

\newcount\Comments  %
\usepackage{color}
\definecolor{darkred}{rgb}{0.9,0,0}
\definecolor{darkgreen}{rgb}{0,0.5,0}
\definecolor{darkblue}{rgb}{0,0,0.7}
\definecolor{purple}{rgb}{.6, 0,.6}
\definecolor{orange}{rgb}{1.0,0.64,0}
\definecolor{mypurple}{RGB}{114,47,238}
\definecolor{DarkBlue}{RGB}{72, 116, 203}
\definecolor{lightcyan}{RGB}{230,250,250}
\newcommand{\kibitz}[2]{\ifnum\Comments=1\textcolor{#1}{#2}\fi}
\newcommand{\modelname}{\textbf{\texttt{SIM1}}}
\newcommand{\modelletter}{\textbf{\texttt{SIM1}}}
\title{\modelname{}: Physics-Aligned Simulator as Zero-Shot Data Scaler in Deformable Worlds}
\correspondingauthor{ Contributions and emails of all authors in \Cref{sec:authors}.}
\renewcommand{\today}{}
\author[*,$\dagger$,1]{Yunsong Zhou}
\author[*,1,2]{Hangxu Liu}
\author[*,1]{Xuekun Jiang}
\author[*,1]{Xing Shen}
\author[1]{\\Yuanzhen Zhou}
\author[1,3]{Hui Wang}
\author[1]{Baole Fang}
\author[1,4]{Yang Tian}
\author[1]{Mulin Yu}
\author[1]{Qiaojun Yu}
\author[1]{\\Li Ma}
\author[1]{Hengjie Li}
\author[1]{Hanqing Wang}
\author[1]{Jia Zeng}
\author[$\dagger$,1]{Jiangmiao Pang}

\affil[1]{Shanghai AI Lab}
\affil[2]{Fudan University}
\affil[3]{Shanghai Jiao Tong University}
\affil[4]{Peking~University}

\begin{document}

\begin{abstract}
Robotic manipulation with deformable objects represents a data-intensive regime in embodied learning, where shape, contact, and topology co-evolve in ways that far exceed the variability of rigids. Although simulation promises relief from the cost of real-world data acquisition, prevailing sim-to-real pipelines remain rooted in rigid-body abstractions, producing mismatched geometry, fragile soft dynamics, and motion primitives poorly suited for cloth interaction. 
We posit that simulation fails not for being synthetic, but for being ungrounded.
To address this, we introduce \modelname{}, a physics-aligned real-to-sim-to-real data engine that grounds simulation in the physical world. Given limited demonstrations, the system digitizes scenes into metric-consistent twins, calibrates deformable dynamics through elastic modeling, and expands behaviors via diffusion-based trajectory generation with quality filtering. 
This pipeline transforms sparse observations into scaled synthetic supervision with near-demonstration fidelity.
Experiments show that policies trained on purely synthetic data achieve parity with real-data baselines at a 1:15 equivalence ratio, while delivering 90\% zero-shot success and 50\% generalization gains in real-world deployment. 
These results validate physics-aligned simulation as scalable supervision for deformable manipulation and a practical pathway for data-efficient policy learning.

\vspace{1em}

\links{
  \link{code}{Code}{https://github.com/InternRobotics/SIM1}, 
  % \link{model}{Model}{https://huggingface.co/InternRobotics/InternVLA-N1}, 
  \link{data}{Data}{https://huggingface.co/datasets/InternRobotics/Sim1-Datset}, 
  \link{homepage}{Homepage}{https://internrobotics.github.io/sim1.github.io/}, 
  \link{demo}{Demo}{https://sim1-demo.intern-robotics.com/} 
}
\vspace{0.5cm}
\end{abstract}

\maketitle

\begin{figure}[h!]
% \centering
\includegraphics[width=.95\linewidth]{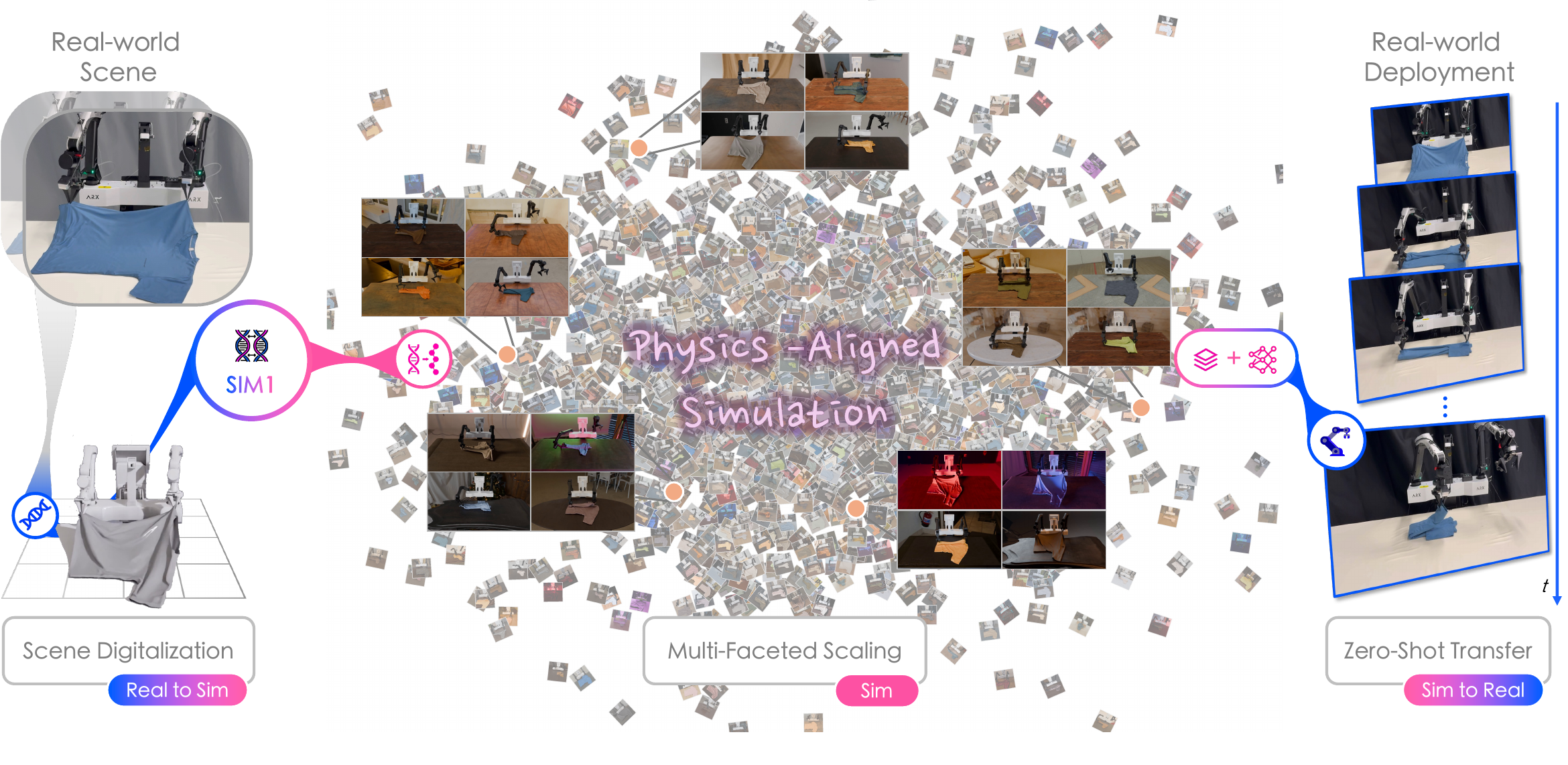}
\caption{\textbf{\modelname{}} pioneers \textit{real-to-sim-to-real} data generation for deformable manipulation.
It constructs \textcolor{mypurple}{sim}ulation data whose deployment behavior 
% is \textcolor{mypurple}{same} as real,
is the \textcolor{mypurple}{same one} as reality,
enabling zero-shot transfer and scalable performance on physical robots.} %
\vspace{-0.5cm}
\label{fig:teaser}
\end{figure} 
\section{Introduction}
\label{sec:intro}

Dexterous manipulation of deformable objects, notably garments, underpins a wide spectrum of daily human activities and remains a frontier for robotics.
Recent advances~\citep{black2024pi_0,cai2026internvla,intelligence2025pi05visionlanguageactionmodelopenworld,bjorck2025gr00t} reveal a clear scaling trend whereby expanding real-robot datasets consistently improves performance, reinforcing the growing consensus on data scaling as a primary driver of embodied learning.
While 
this paradigm shows promise
% promising
in rigid-object settings, deformable manipulation intensifies the hunger for data, as its evolving geometry and contact-rich dynamics demand substantially broader state and visual coverage.
Yet, acquiring real-world manipulation data at scale remains prohibitively expensive.
% poses a substantial obstacle.
% Scaling such datasets 
Such efforts
demands skilled operators, specialized hardware, and extensive human labor, placing practical limits on dataset scale.
% rendering large-scale data collection impractical for most research teams. 
Consequently, vision-language-action models trained on limited deformable-interaction data exhibit constrained generalization.

In the field, sim-to-real (S2R) synthetic data generation has emerged as a compelling strategy for scaling manipulation data~\citep{ gu2023maniskill2,ye2025dex1b,he2025viral,xue2025openingsimtorealdoorhumanoid,deng2025graspvlagraspingfoundationmodel}. 
In rigid-object domains, increasing synthetic diversity often translates into measurable real-world gains, supported by rich asset libraries and automated generation pipelines~\citep{tian2025interndata,chen2025robotwin,robocasa365}.
However, this paradigm breaks down in deformable manipulation.
\textit{Simulated scenes} are often weakly aligned, or not aligned at all, with real environments, relying on coarse calibration or manually constructed assets without metric fidelity~\citep{Gao_2025_CVPR,wang2024grutopia,pmlr-v205-li23a,lu2024garmentlab}.
Such geometric inconsistencies are amplified under soft-body deformation.
\textit{Physics engines} are predominantly optimized for rigid-body dynamics, producing unstable or inaccurate responses for cloth and other soft materials~\citep{chen2025robotwin,zhang2024dexgraspnet,deng2025graspvlagraspingfoundationmodel}.
\textit{Behavior generation} follows rigid-object paradigms based on grasp-point and simple pick-and-place primitives, rendering these strategies inadequate for garment manipulation~\citep{robosplat}.
As a result, synthetic data serves primarily as pre-training signals, while reliable performance still depends on real-world post-training.

We contend that the first principle of simulation is grounding; scaling becomes valuable only once simulated physics correspond to reality.
Real-to-sim (R2S) alignment is thus foundational for deformable manipulation, prioritizing correspondence between simulated and physical dynamics over superficial realism or asset import~\citep{tian2025interndata,yin2026geniesim30}.
This motivates a shift from post-hoc adaptation to alignment-first simulation, raising a central question:
\textit{how can simulation be grounded so that synthetic data is a reliable substrate for real-robot deployment?}

To address this, we advocate that a real-to-sim-to-real (R2S2R) paradigm offers a principled path toward real-equivalent simulation by aligning each stage with the physical world, thereby enabling direct transfer to real robots~\citep{torne2024rialto}.
As depicted in \Cref{fig:teaser},
we present \modelname{}, the \textit{first} physics-aligned simulation pipeline for deformable manipulation that digitalizes real scenes (R2S) and generates scalable, unbiased synthetic data consistent with real-world settings (S2R).
By expanding limited demonstrations into diverse training data, \modelname{} enables policies that transfer directly to real-robot deployment without additional tuning.

Specific designs operationalize the R2S2R paradigm across three complementary alignment stages.
For \textit{geometric alignment}, high-precision 3D scans are reconstructed into metric-accurate, textured meshes, producing simulation-ready digital representations of real-world scenes.
For \textit{dynamical alignment}, a stabilized soft-body solver~\citep{Giles2025} enforces physically consistent elastic and bending responses while suppressing excessive deformation, thereby enabling realistic interaction modeling. 
A coupled simulation infrastructure maps robot operations to simulation, supporting parameter calibration and stable soft-body manipulation.
For \textit{movement alignment}, deformable manipulation trajectories are synthesized through structured two-stage planning that decouples interaction from motion. Diffusion-based trajectory generation models human-like behavior~\citep{NEURIPS2024_2aee1c41}, while automatic filtering and appearance randomization~\citep{8202133} enhance diversity and robustness.
These components transform simulation into a real-equivalent data source for scalable generation to real robots.

We summarize our contributions as follows: 
1) We introduce \modelname{}, which minimizes the sim-to-real gap through a physics-aligned R2S2R paradigm, enabling synthetic data to serve as high-fidelity training data for direct deployment in deformable manipulation.
2) We enhance simulation fidelity and data utility through metric-accurate scene digitization, a deformation-stabilized solver with physics-based calibration, and a diffusion-based motion framework coupled with filtering to generate high-quality manipulation data.
3) Experiments on $\pi_{0.5}$ and $\pi_{0}$ achieve zero-shot success rates of 90\% and 76\%, with generalization gains of +50\% and +56\% over real-data baselines. Furthermore, 15 synthetic samples from \modelname{} provide training value comparable to a real demonstration, validating its effectiveness as a data scaler for deformable manipulation.

\section{Related Work}

\noindent \textbf{Data scaling and manipulation datasets.}  
Scaling real-robot data improves vision-language-action models (VLAs)~\citep{black2024pi_0,intelligence2025pi05visionlanguageactionmodelopenworld,bjorck2025gr00t,cai2026internvla,chen2025internvla,cheang2024gr2generativevideolanguageactionmodel,kim24openvla,yang2026rise,li2026forcevla2,sima2026kai0,bu2025univla,bu2025agibot_iros,zheng2026egoscale,gao2026dreamdojo,ye2026world,tian2025simscale}. Large-scale datasets such as Open X-Embodiment~\citep{open_x_embodiment_rt_x_2023}, DROID~\citep{khazatsky2024droid}, and BridgeData v2~\citep{pmlr-v229-walke23a} advance rigid and articulated manipulation, while simulation datasets (\textit{e.g.}, ManiSkill2~\citep{gu2023maniskill2}, RoboCasa~\citep{robocasa365}, Genie Sim 3.0~\citep{yin2026geniesim30}) extend data coverage for rigid scenarios. However, deformable manipulation remains underrepresented: soft-body dynamics introduce continuous shape variation and contact-rich behaviors that are more data-hungry than rigid-object tasks. Our work addresses this gap by using simulation to synthesize scalable deformable demonstrations, expanding training data specifically for soft-object manipulation.

\noindent \textbf{Simulation-to-real synthetic data generation.}  
Recent methods generate large-scale demonstrations via transformation~\citep{mandlekar2023mimicgen,11127809}, R2S alignment~\citep{torne2024rialto,huang2026soma}, and physical twin reconstruction~\citep{jiang2025phystwin}, achieving strong results in rigid manipulation and emerging capabilities for humanoid~\citep{xue2025openingsimtorealdoorhumanoid,he2025viral} and deformable tasks~\citep{yu2025right}. Nevertheless, S2R degradation persists because simulated dynamics are not fully aligned with real-world physics. Our R2S2R framework addresses this by grounding simulation in physical reality through mesh-level alignment, physics-faithful dynamics solving and human-like trajectory generation, producing synthetic data that supports zero-shot transfer to real robots.

\noindent \textbf{Real-to-simulation asset digitization.}  
The digitization of deformable objects employs MPM~\citep{chenhu2026empm}, spring-mass models~\citep{jiang2025phystwin}, and platforms such as GarmentLab~\citep{lu2024garmentlab}, which integrate multiple physics engines. However, solver accuracy remains limited: VBD suffers from unrealistic stretching~\citep{10.1145/3658179}, while PBD and FEM involve trade-offs between accuracy and efficiency~\citep{MULLER2007109,10.1145/566654.566623}. Recent solvers~\citep{Giles2025} improve strain limiting but remain isolated from broader pipelines. 
While existing solvers achieve accurate offline deformation through particle-state optimization, they are not designed for the real-time requirements of embodied manipulation where rigid–soft interaction must be updated dynamically during control. 
Our solver and calibration infrastructure support online rigid–soft coupling with stable deformation dynamics, enabling high-fidelity simulation and data generation for deformable manipulation.

\begin{figure}[t]
  \centering
  \includegraphics[width=\linewidth]{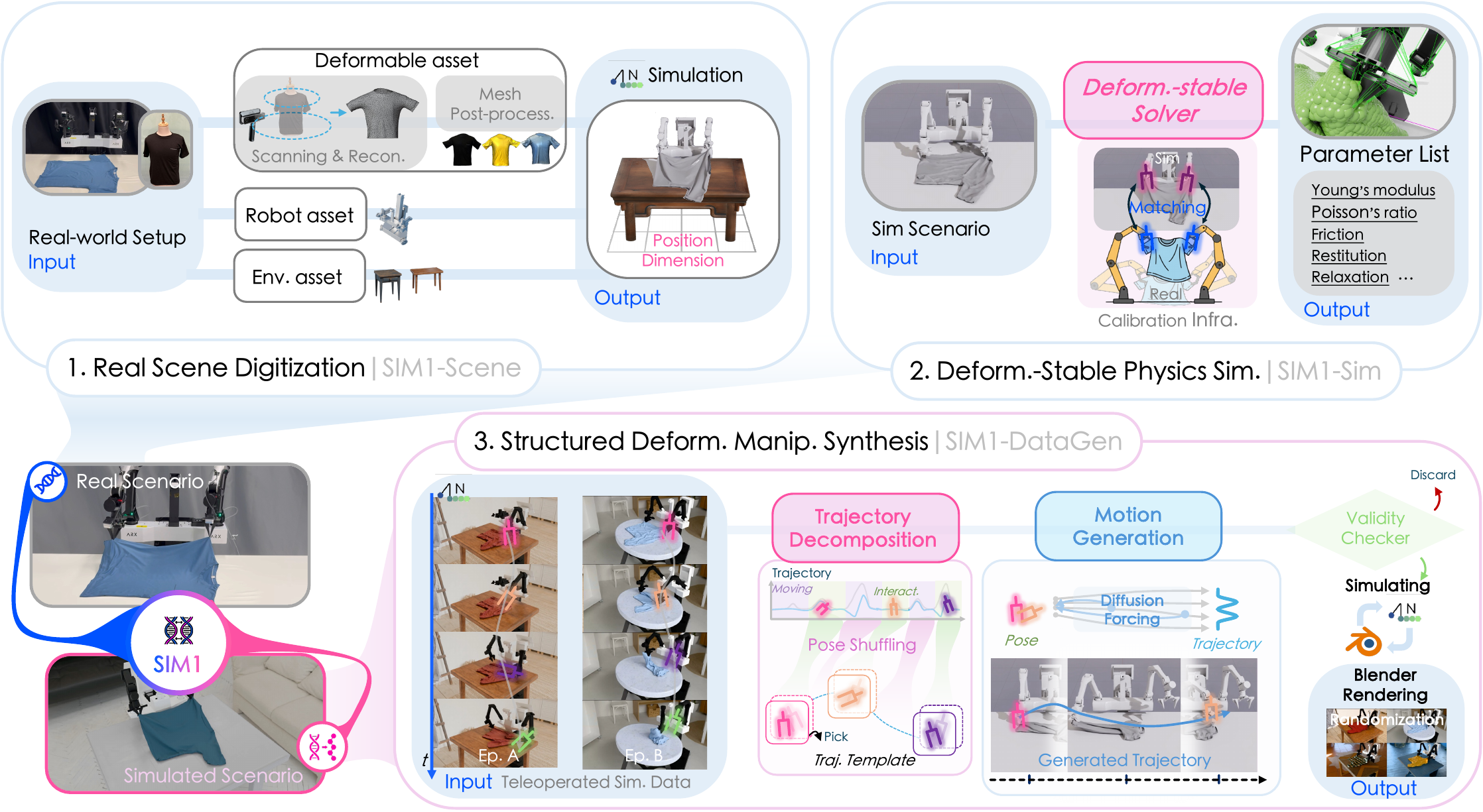}
  \caption{\textbf{Framework of \modelname{}.}
  \textbf{(1)} Real-world objects are reconstructed into metric-accurate, textured simulation assets; 
  \textbf{(2)} They are then executed within a deformation-stable simulation framework calibrated through real-to-sim behavior matching. 
  \textbf{(3)} Upon physical alignment, diverse manipulation trajectories are synthesized via structured subtask decomposition and diffusion-based motion generation, and rendered with appearance randomization to produce real-equivalent synthetic training data.}
  \label{fig:framework}
\end{figure}

\section{\modelname{} Framework}
\label{sec:framework}

\modelname{} adopts the real-to-sim-to-real (R2S2R) paradigm to bridge geometry, dynamics, and motion across stages. 
This approach addresses the asymmetry of one-way simulation or reconstruction methods and enables synthetic data to function as a real-equivalent substrate for robot learning (\Cref{fig:framework}).
High-precision scans (\textit{e.g.}, garment meshes) and object imports (\textit{e.g.}, URDFs, environment assets) are converted into metric-accurate digital scenes, providing simulation-ready geometric configurations (\Cref{sec:real_world_extraction}).
Within the aligned simulator, a deformation-stabilized solver and parameter calibration infrastructure reproduce realistic dynamics, enabling interactive soft-body manipulation with calibrated physics (\Cref{sec:real_to_sim_matching}).
Demonstration data from teleoperated simulation are first decomposed into motion segments and subsequently synthesized via diffusion, with visual randomization used to generate scaled training data that enhances generalization (\Cref{sec:sim_scene_generation}).

\subsection{\modelletter{}-Scene: Real Scene Digitization}
\label{sec:real_world_extraction}

The first stage establishes metric-accurate geometric alignment as a prerequisite for closing the sim-real gap.
Since even minor discrepancies in shape, scale, or spatial configuration can propagate into dynamic and contact errors, we construct static simulation assets that faithfully reproduce their real-world counterparts. 
In particular, recovering high-quality meshes for deformable clothing is complicated by fine wrinkles and intricate topology. 
We address this challenge by combining high-precision 3D scanning with dedicated mesh post-processing to obtain textured and dimensionally accurate models. 
Robots and static environments are incorporated using calibrated URDF imports and asset libraries. 
The following sections detail the digitization for each asset type.

\noindent \textbf{Deformable assets.}
For garments, we employ a professional 3D scanner (EinScan Rigil Pro) to capture high-fidelity meshes and textures.
The garment is mounted on a mannequin to maintain its natural shape during scanning. 
Multi-view RGB images and LiDAR scans are captured and fused to generate a dense point cloud.
As the scan includes the mannequin, we perform a manual segmentation step to remove mannequin points, retaining only the garment.
The resulting point cloud is then processed through surface refinement (\textit{e.g.}, Poisson reconstruction~\citep{10.1145/2487228.2487237}) followed by mesh post-processing, including hole filling, smoothing, and remeshing to obtain a clean, watertight mesh suitable for simulation.
Texture is mapped from the RGB images onto the static mesh, resulting in a textured OBJ asset.
This process yields a geometric replica of the real garment with submillimeter precision.

\noindent \textbf{Robot assets.}
The robot used in this study is the ARX ACONE robot, a bimanual platform designed for dexterous manipulation tasks.
Its kinematic structure, joint limits, collision geometries, and visual meshes are defined in a URDF file generated from CAD models (\textit{e.g.}, SolidWorks) provided by the manufacturer.
We directly import this URDF into our simulation environment, ensuring that each arm's degrees of freedom and workspace exactly match the real hardware. No additional scanning is required; however, we verify dimensional accuracy and adjust the root transform to align with the world coordinate frame.
The two arms are calibrated relative to each other to preserve correct bimanual coordination.

\noindent \textbf{Environment assets.}
Static objects in the environment (\textit{i.e.}, diverse tables) are obtained from publicly available 3D model repositories or created manually.
These assets are imported as mesh files and placed in the scene at positions and orientations that replicate the real-world setup.
Dimensions are scaled according to real-world measurements to maintain physical consistency.

\begin{figure}[t]
\centering
\includegraphics[width=0.99\linewidth]{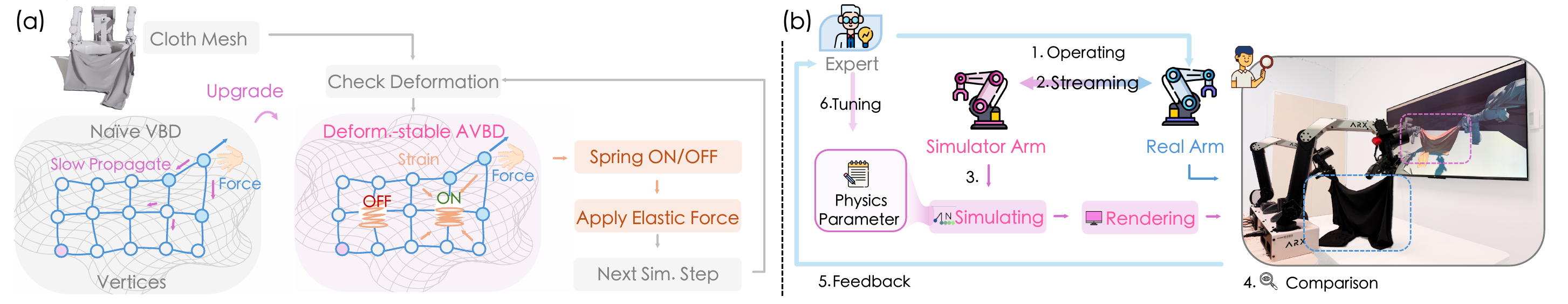}
\caption{\textbf{Paradigm of deformation-stable physics simulation.}
\textbf{(a)} After naive VBD~\citep{10.1145/3658179} updates under external forces, edge deformation is monitored and virtual elastic constraints are activated when stretch exceeds a threshold, injecting strain forces that accelerate convergence toward physically plausible cloth configurations. 
\textbf{(b)} A bidirectionally synchronized simulation infrastructure replaces identical dual-arm executions in simulation and aligns deformation behaviors through visual calibration.
}
\label{fig:solver}
\end{figure}

\subsection{\modelletter{}-Sim: Deformation-Stable Physics Simulation}
\label{sec:real_to_sim_matching}

Achieving reliable S2R transfer for deformable manipulation requires physically consistent rigid–soft coupling, a setting that remains poorly supported by existing simulation engines. 
Modern physics solvers are typically optimized for either rigid-body (widely used in robotics~\citep{wang2024grutopia,pmlr-v205-li23a,chen2025robotwin}) or soft-body (largely studied in graphics~\citep{stiffgipc2025,10.1145/566654.566623,li2020incremental,10.1145/3340258}), but not their interaction. 
In soft-body solvers, deformation is resolved through iterative updates of particle vertex states, where contact forces propagate gradually across the mesh (\Cref{fig:solver} (a)). 
While accurate deformation can emerge with sufficient solver iterations in offline simulation, embodied manipulation requires cloth states to update in lockstep with robot motion during online teleoperation.
Limited force propagation and delayed strain updates, therefore, lead to excessive stretching, unstable contact, and misalignment with real worlds (\Cref{sec:avbd_implementation}). 
To close this gap, we design a deformation-stable solver that constrains strain evolution during contact, enabling stable soft-body dynamics under real-time interaction, 
and establish a calibration infrastructure that aligns simulated and real behaviors under identical control inputs.

\noindent \textbf{Stabilized solver via Augmented Vertex Block Descent.}
We develop a deformation-stable solver inspired by the Augmented Vertex Block Descent (AVBD) formulation~\citep{Giles2025}, extending the Newton–VBD solver~\citep{10.1145/3658179}.
Instead of relying solely on energy minimization to update vertex positions, our solver introduces explicit strain constraints between vertices during optimization.
These constraints act as adaptive elastic links that rapidly propagate forces across the cloth mesh, allowing vertex motion to remain consistent with gripper dynamics.
This design stabilizes rigid–soft interaction while preserving the efficiency required for large-scale simulation data generation.

\noindent \textbf{Solver workflow.}
As illustrated in \Cref{fig:solver} (a), the solver operates on a cloth mesh whose vertices represent particles connected by edges.
When external forces are applied (\textit{e.g.}, robot contact), vertex positions are first updated through the standard Newton–VBD iteration.
After each update, we examine the deformation of every mesh edge.
If the distance between two connected vertices exceeds a predefined stretch threshold, a virtual elastic constraint is activated.
This constraint introduces an additional tensile force between the vertices, which is injected into the VBD optimization and accelerates convergence toward physically plausible configurations.
The procedure iterates until vertex positions stabilize.

\noindent \textbf{Strain constraint.}
Consider an edge $\mathbf{e}$ connecting vertices $i$ and $j$ with rest length $l_0$.
Let $\mathbf{e}_i, \mathbf{e}_j \in \mathbb{R}^3$ denote their current positions.
We define a maximum stretch ratio $\xi>0$ such that the edge length should not exceed $(1+\xi)l_0$.
The strain constraint is therefore:
\begin{equation}
C(\mathbf{e})
= \|\mathbf{e}_i - \mathbf{e}_j\| - (1+\xi)l_0
\le 0.
\label{eq:constraint}
\end{equation}
The constraint becomes active when $C(\mathbf{e})>0$, indicating that the local deformation exceeds the allowable limit.

\noindent \textbf{Constraint energy formulation.}
When activated, we apply an additional energy term that penalizes excessive stretch:
\begin{equation}
E^{(n)}_{\text{strain}}(\mathbf{e}) =
\begin{cases}
\frac{1}{2} k^{(n)} C(\mathbf{e})^2
+ \lambda^{(n)} C(\mathbf{e}),
& C(\mathbf{e})>0, \\
0, & \text{otherwise},
\end{cases}
\label{eq:energy}
\end{equation}
where $k^{(n)}$ is the penalty stiffness parameter at Newton iteration $n$, and $\lambda^{(n)}$ is the Lagrange multiplier accumulating constraint forces.
The gradient of $E^{(n)}_{\text{strain}}$ are added to the Newton system and solved within the VBD vertex updates.
Intuitively, this term acts as a virtual spring that activates only when excessive stretching occurs, injecting corrective forces that guide vertices toward physically consistent configurations.

\noindent \textbf{Parameter update.}
After each Newton iteration, the penalty stiffness and variables are updated to progressively enforce the strain constraint:
\begin{equation}
%\[
k^{(n+1)}=
\min\!\left(k_{\max},\;k^{(n)}+\beta|C(\mathbf{e})|\right),
\quad
\lambda^{(n+1)}=
\lambda^{(n)}+k^{(n)}C(\mathbf{e}),
%\]
\end{equation}
where $k_{\max}$ is the maximum stiffness and $\beta$ controls the ramping rate.
This update integrates correction forces over iterations, preventing explosive stretching and improving deformation stability during contact-rich manipulation.

\noindent \textbf{Simulation infrastructure for parameter calibration.}
Physical parameters list $\Theta = \{\rho, E, \nu, \mu, \eta, \zeta\}$ (density, Young's modulus, Poisson's ratio, friction, restitution, relaxation) cannot be recovered to their true physical values through direct optimization. 
Instead, we calibrate simulation by aligning its behavior with real observations.
\Cref{fig:solver} (b) shows the calibration pipeline following a bidirectional workflow. 
An expert operates the dual-arm robot to execute representative manipulation sequences. 
The robot’s joint states are streamed to the simulator so that the simulated twin reproduces identical motions. 
The renderings are visually compared with real executions, allowing experts to assess discrepancies in draping, folding, and contact behavior.
Based on this visual feedback, parameters $\Theta$ are iteratively adjusted until simulated interactions exhibit behavior that matches the real system at an operational level. 
This process does not guarantee recovery of true physical parameters, but it establishes behavioral consistency: simulated scenes respond to manipulation in ways that closely resemble reality.

\subsection{\modelletter{}-DataGen: Structured Deformable Manipulation Synthesis}
\label{sec:sim_scene_generation}

Rigid-object manipulation can often rely on trajectory slicing and recomposition of motion primitives~\citep{mandlekar2023mimicgen,11127809}. 
For deformable objects, however, contact dynamics are state-dependent and highly non-deterministic: valid grasp locations cannot be reliably detected, and naive slicing breaks interaction fidelity. 
To address this, we generalize trajectory decomposition by decoupling interaction from motion (\Cref{fig:framework} right-bottom). 
Grasp configurations are directly reused from expert demonstrations; selection is randomized to increase diversity while preserving the original ordering.
Motion between interactions is synthesized via diffusion-based generation, learning human-like transitions from demonstration data. 
Visual randomization further expands appearance diversity, producing large-scale training data suitable for policy learning (\Cref{sec:appendix_sim_data_gen}).

\noindent \textbf{Trajectory decomposition.}
Given demonstrations $\mathcal{D}=\{\tau_i\}$ collected via teleoperation, each trajectory is segmented into \texttt{interacting} and \texttt{moving} regions. 
\texttt{Interacting} segments encode stable grasp configurations and contact states; these are preserved and pooled without modification. 
To generate new demonstrations, we randomly pick segments from this pool, reusing templated grasp poses while varying sequence order and task context. 
This shuffling operation expands behavioral diversity while maintaining physically valid interactions, addressing the lack of reliable grasp detection under deformation.

\noindent \textbf{Diffusion-based motion generation.}
Between consecutive \texttt{interacting} segments, we generate smooth transitions by treating trajectory synthesis as sequence completion. Given picked boundary poses $(\mathbf{p}_s,\mathbf{p}_e)$ and robot history $\mathbf{h}$, the model predicts intermediate motions that satisfy physical and kinematic consistency. We employ conditional diffusion forcing~\citep{NEURIPS2024_2aee1c41}, where a transformer sequence model reconstructs trajectories from partially corrupted tokens. The diffusion process applies stochastic masking with noise level $\mathbf{k}$, and learning proceeds by recovering the original sequence from soft-corrupted inputs. Formally,
\begin{equation} 
\forall ~\mathbf{k} \in (0,1]^{\mathcal{T}},\ \underset{\theta}{\text{min}}\ \mathbb{E}\big\| \mathbf{\epsilon} - \epsilon_{\theta}\big(g(\mathbf{x}^0,\mathbf{k});\mathbf{h},\mathbf{p}_s, \mathbf{p}_e,\mathbf{k}\big) \big\|_2^2, \end{equation}
where $g$ denotes the corruption function that applies noise to the clean sequence $\mathbf{x}^0$ according to moise $\mathbf{k}$, and $\epsilon_{\theta}$ is a transformer predicting the noise residual conditioned on history and keyposes.
This formulation enables physically consistent motion synthesis that bridges interaction segments while preserving realism.

\noindent \textbf{Validity checking.}
Even with optimization, rigid–soft interaction is inherently uncertain: gripper contacts with deformable objects can produce penetration or adhesion, yielding physically implausible trajectories.

To filter such failures, we first perform lightweight state-based filtering using garment particle states.
From simulation-derived positive and negative trajectories, we leverage vibe coding to synthesize threshold rules over particle statistics, defining admissible regions that favor positive states and exclude negative ones. 
This step efficiently prunes invalid samples early.
We then train a binary video discriminator $D$ on head-view RGB observations $\mathbf{V}$ to distinguish valid demonstrations from low-quality samples. Negative cases arise naturally during simulation, avoiding manual annotation. A ResNet-18 feature extractor~\citep{He_2016_CVPR} and Transformer encoder~\citep{NIPS2017_3f5ee243} aggregate temporal information and output a validity score $s=D(\mathbf{V})$. Trajectories with $s>\tau_{\text{disc}}$ are retained; others are discarded. This lightweight filtering removes implausible demonstrations without costly physics-based validation.

\noindent \textbf{Visual randomization.}
Valid trajectories are rendered in Blender~\citep{blender} with appearance randomization of materials, lighting, and camera parameters. Multiple variations are generated per trajectory using cycle path tracing to produce photorealistic RGB images synchronized with trajectory timestamps. The final dataset combines rendered observations with robot states and actions in the LeRobot format~\citep{lerobot2024} for imitation learning.

\begin{figure}[t]
\centering
\includegraphics[width=0.99\linewidth]{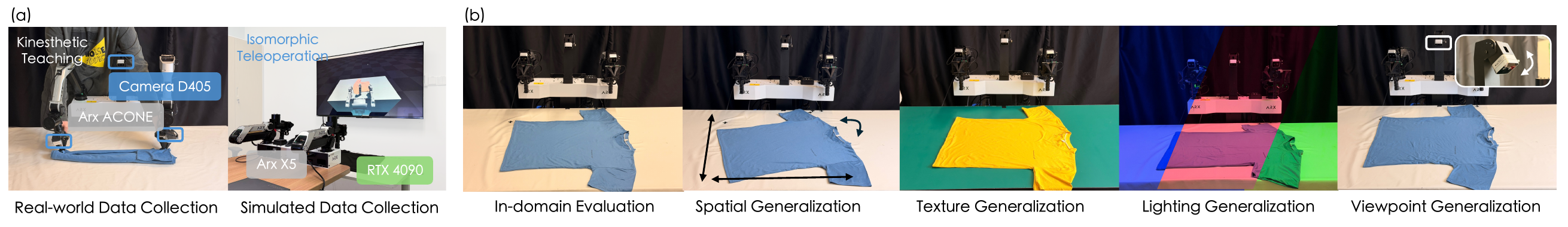}
\caption{\textbf{Illustration of data collection and evaluation.} \textbf{(a)} Real-world and simulated data collection via kinesthetic teaching and isomorphic teleoperation on Arx ACONE and Arx X5. 
\textbf{(b)} Domain settings for in-domain and out-of-domain evaluation in real-world experiments.
Representative long-horizon T-shirt folding task (over 20 seconds) illustrating complex sequential manipulation capabilities.
}
\label{fig:experiment}
\end{figure}

\begin{figure}[t]
    \centering
    \includegraphics[width=0.99\linewidth]{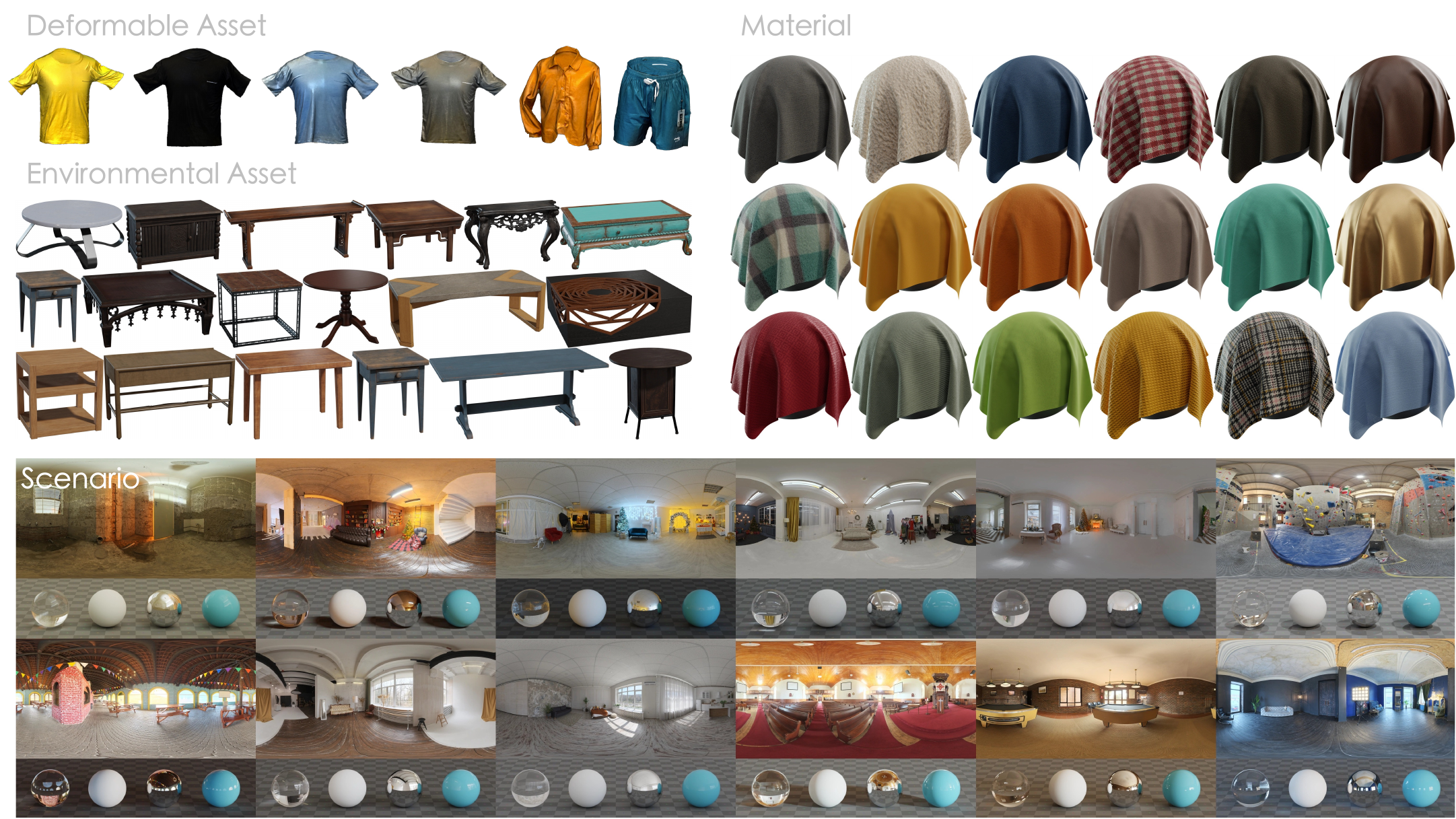}
    \caption{\textbf{Illustration of assets used in data generation.} Scanned deformable assets and open-source environmental assets used in simulation (top-left).
    Diverse garment textures for appearance variation (top-right).
    Room-scale environments with randomized layouts and lighting for scene-level randomization (bottom).}
    \label{fig:asset}
\end{figure}

\begin{figure}[t]
    \centering
    \includegraphics[width=0.99\linewidth]{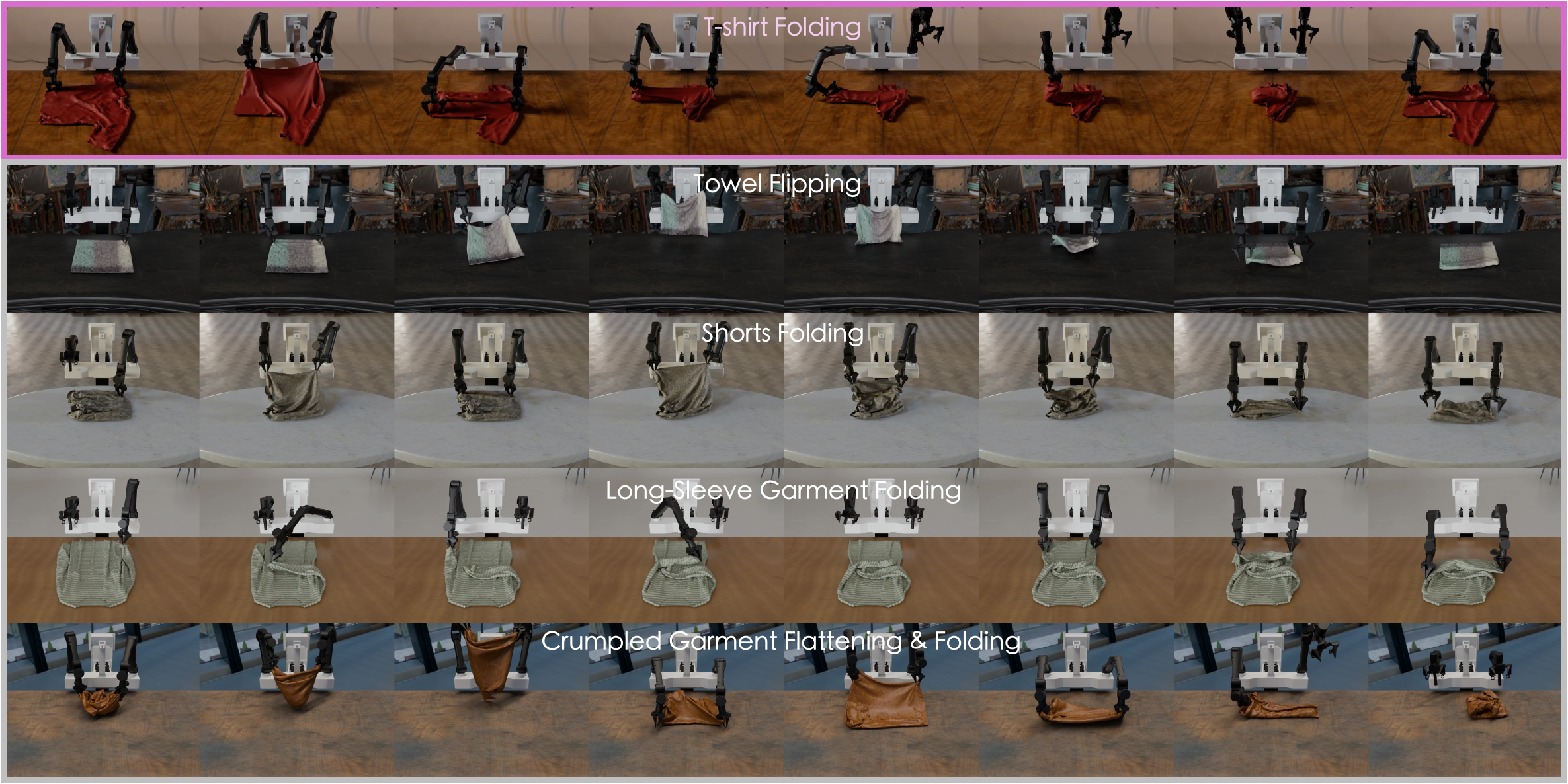}
    \caption{\textbf{Visualization of generated data across garments and tasks.} The pink box highlights the representative T-shirt folding task used as the real-world benchmark in this work.
    The gray box shows additional garments and manipulation tasks generated in simulation, including folding, flipping, and flattening, illustrating the broader task coverage enabled by our framework beyond the single benchmark task.}
    \label{fig:data}
\end{figure}

\section{Experiments}
We evaluate whether a physics-grounded simulation pipeline can serve as a scalable and reliable substitute for real-world data in structured deformable manipulation. 
Our experiments focus on S2R transfer, cross-domain generalization, and data scaling efficiency under purely simulation-trained policies.
Specifically, we investigate the following questions.
\textbf{Q1}: Can models trained solely in simulation achieve performance comparable to real-data-trained counterparts?
\textbf{Q2}: Does simulation-induced diversity improve out-of-domain robustness beyond real-world data?
\textbf{Q3}: Does synthetic data enable more efficient performance gains through scaling than real-world collection?

\begin{figure}[t]
    \centering
    \includegraphics[width=0.99\linewidth]{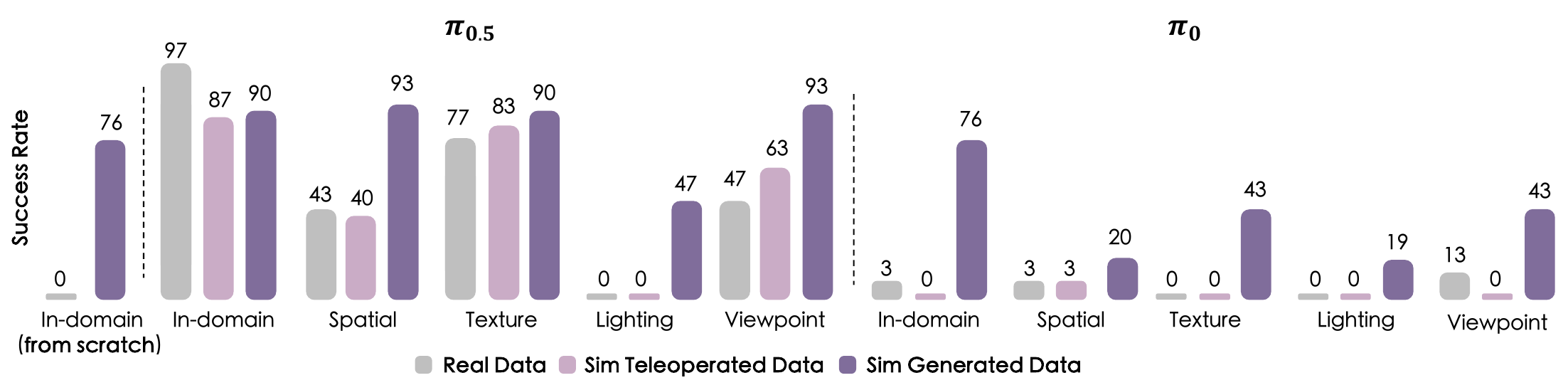}
    \caption{\textbf{In-domain and out-of-domain evaluation.}
    Policies are trained with real data, collected simulation data, or simulator-generated data.
    Groups: $\pi_{0.5}$ trained from scratch, $\pi_{0.5}$ post-trained, and $\pi_0$ post-trained. Simulated data matches real episodes under equal budgets and surpasses them when scaled, especially under domain shifts.}
    \label{fig:main_number}
\end{figure}

\begin{figure}[!t]
    \centering
    \includegraphics[width=0.99\linewidth]{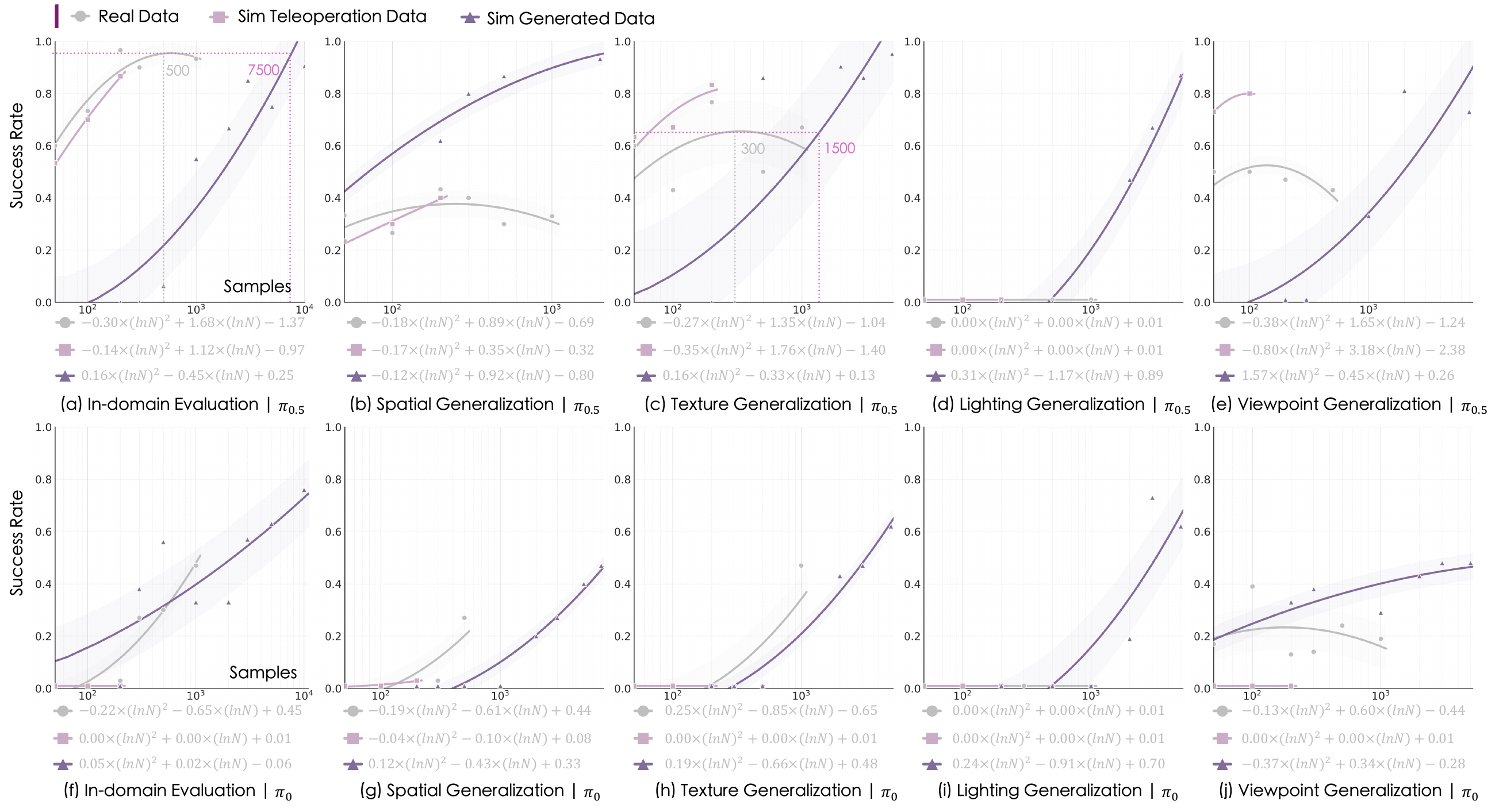}
    \caption{\textbf{Curves of performance versus data scale.}
    Synthetic data scaling improves performance and can surpass real-data-only training. Dashed lines denote the equivalence points where $M$ synthetic samples match one real sample at saturation.}
    \label{fig:main_result}
\end{figure}

\begin{figure}[!t]
    \centering
    \includegraphics[width=0.99\linewidth]{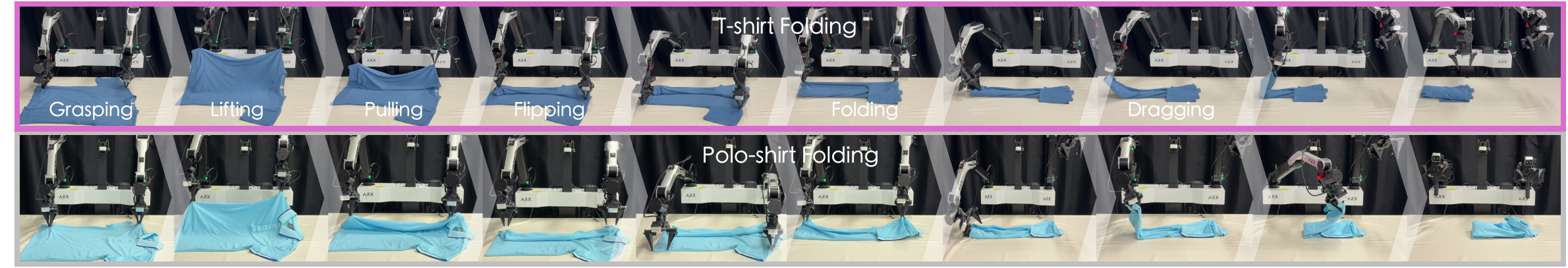}
    \caption{\textbf{Qualitative results of zero-shot sim-to-real transfer.} Top: Real-robot deployment of our representative T-shirt folding task using policies trained on synthetic data, illustrating successful folding from grasping to completion.  
    Bottom: Deployment on highly dissimilar garments with material, shape, and size configurations absent from training and simulation, demonstrating generalization across extreme domain shifts.}
    \label{fig:qualitative_results}
\end{figure}

\begin{figure}[t]
    \centering
    \includegraphics[width=0.99\linewidth]{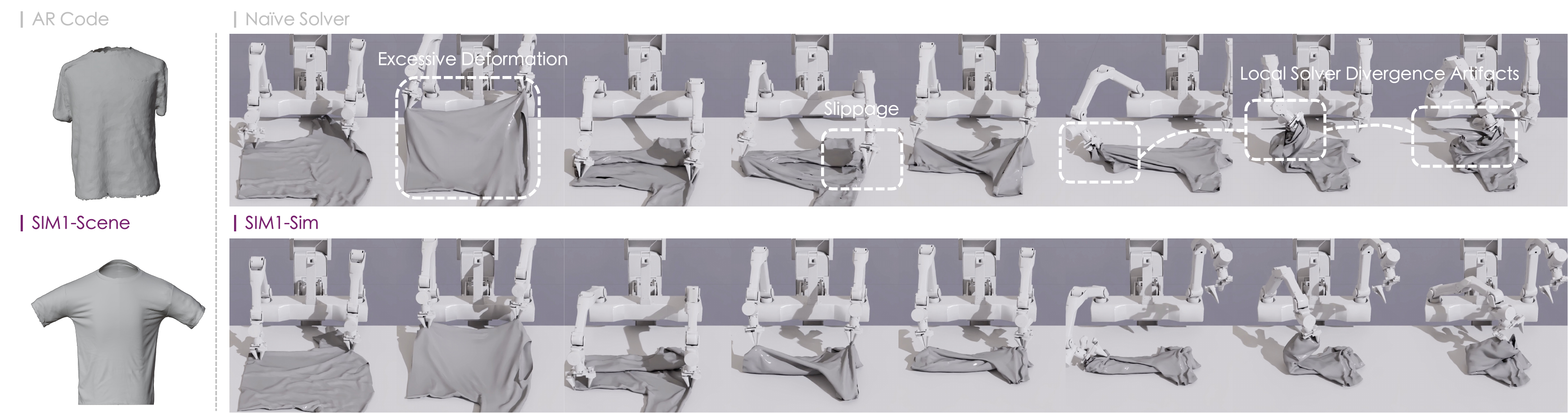}
    \caption{\textbf{Scene digitization and solver stability.} 
    Our pipeline produces detailed geometry suitable for simulation, whereas marker-based methods (\textit{e.g.}, AR Code) yield coarser reconstructions. Conventional solvers exhibit artifacts in rigid–soft interaction (dashed regions); our solver remains stable and closely matches real-world behavior.}
    \label{fig:solver_exp}
\end{figure}

\subsection{Protocal and Setup}

\noindent \textbf{Data collection.}
Data are collected in both real and simulated environments. 
In \Cref{fig:experiment} (a), we use an ARX ACONE dual-arm platform equipped with parallel-jaw grippers. 
In the real world, we adopt kinesthetic teaching, in which the operator directly guides the robot's end-effectors by hand; in total, 1,000 real trajectories are recorded. 
In simulation, we deploy our Newton-based simulation infrastructure with the deformation-stable AVBD solver on an RTX 4090 workstation, aligning garment geometry, robot kinematics, and camera parameters with the physical setup.
A teleoperator controls two ARX X5 arms, whose motions are mapped to the simulated robot in real time. The operator observes rendered simulator views and performs folding through visual teleoperation. This process yields 200 source simulation demonstrations, which serve as seeds for subsequent synthetic data generation.

\noindent \textbf{Tasks and baselines.}
T-shirt folding is used as a representative benchmark for real-world validation and does \textbf{\textit{not}} constrain the scope of the proposed simulation pipeline, which targets a broader class of deformable manipulation tasks.
We introduce additional manipulation tasks in the following section to further demonstrate the general applicability of our framework.
Policies are trained using either real demonstrations or our synthetic data exclusively,
and evaluated over 30 trials per configuration. 
A trial is considered successful if the garment reaches the target folded configuration without dropping or unfolding.
We conduct both in-domain and generalization evaluations (\Cref{fig:experiment} (b)). 
In-domain experiments use the same spatial layout, garment, table surface, lighting, and camera configuration as training. 
For generalization, we introduce controlled distribution shifts along multiple factors: the garment is randomly translated up to 8\,cm and rotated within $\pm15^\circ$ (spatial); unseen table and garment textures are substituted with corresponding changes in friction properties (texture); illumination direction and intensity are randomized (lighting); and the camera elevation is perturbed within $\pm5^\circ$ (viewpoint).

\noindent \textbf{Synthetic data generation.}
For synthetic data generation, we follow the pipeline in \Cref{sec:appendix_sim_data_gen}. The framework supports diverse asset combinations and environment variations; in this work, we use a representative configuration for real-world validation while demonstrating the general capability of the approach. To enhance diversity during synthesis, the garment is translated within 5\,cm and rotated within $\pm15^\circ$, table and cloth materials are sampled from 17 and 28 types, respectively, and 90 randomized environment combinations are applied with head-camera elevation perturbations of $\pm5^\circ$ (\Cref{fig:asset}). Generated samples are visualized in \Cref{fig:data}.

\subsection{Main Experiment Results}

We evaluate policies trained in simulation and real data under both in-domain and out-of-domain settings (\Cref{fig:main_number}). Real demonstrations and teleoperated simulation data use 200 samples, while simulation-generated data are scaled to 10k samples for in-domain training and 2k samples for out-of-domain evaluation.

\noindent
\textbf{Simulation versus real-data performance (A1).}
Simulation-trained policies achieve performance comparable to real-data-trained counterparts under equal data budgets. 
For the representative $\pi_{0.5}$ setting, real data reach average success $\textbf{97}\%$, while policies trained on sim-teleoperated data achieve $\textbf{87}\%$, a marginal gap of $\textbf{10}\%$. This suggests that physics-aligned simulation provides supervision of considerable fidelity to match real-world training with controlled data volumes.

\noindent
\textbf{Out-of-domain robustness (A2).}
Simulation-induced diversity yields substantial generalization gains under domain shifts. In spatial shifts, texture variation, and lighting perturbations, simulation-trained policies outperform real-data-trained baselines by $\textbf{50}\%$, $\textbf{13}\%$, and $\textbf{47}\%$, respectively. These improvements indicate that simulation enables broader coverage of variations than limited real-world demonstrations, enhancing robustness beyond the training distribution.

\noindent
\textbf{Pretraining confound analysis.}
To isolate the effect of pretrained priors, we evaluate task learning under de novo initialization with no reliance on preexisting manipulation knowledge. 
The real-data baseline ($\pi_{0.5}$ trained from scratch) fails completely (\textbf{0}\% success), indicating that limited real demonstrations alone do not enable deformable manipulation. In contrast, the synthetic training pipeline achieves strong task acquisition (\textbf{76}\%) under the same from-scratch condition, demonstrating that performance gains originate from generated data rather than pretrained skills. These results dispel the possibility that task success is driven by prior knowledge and confirm the effectiveness of synthetic supervision.

\noindent\textbf{Scaling analysis (A3).}
\Cref{fig:main_result} examines whether synthetic data enables more efficient performance gains through scaling than real-world collection. As data volume increases, the success rates of $\pi_{0.5}$ and $\pi_{0}$ improve steadily, following the fitted scaling curves.
Simulation demonstrates favorable scaling behavior compared to real-world data. For the representative $\pi_{0.5}$ model under in-domain evaluation, one real demonstration provides comparable benefit to approximately \textbf{15} synthetic samples. In a representative out-of-domain setting (texture generalization), the equivalence shifts to roughly \textbf{5} synthetic samples per real sample.

\noindent \textbf{Findings.}
(1) Synthetic data is weak in extremely low-data regimes but scales more effectively than real data as volume increases. Performance grows rapidly with additional simulation samples and eventually surpasses real-data training, while real-data gains saturate due to limited diversity.  
(2) $\pi_{0.5}$ outperforms $\pi_{0}$ under fixed budgets because of richer pretraining or a more expressive state representation. $\pi_{0}$ requires greater data volume to achieve similar performance, consistent with lower data efficiency rather than fundamental task limitations.

\noindent\textbf{Qualitative Results.}
\Cref{fig:qualitative_results} shows real-world deployments of policies trained purely on synthetic data. 
The policy successfully performs garment folding on a physical robot without any real demonstrations.
It further generalizes to an unseen polo shirt with different material, texture, and geometry, where the real-data baseline achieves $\textbf{20}\%$ success while ours reaches $\textbf{70}\%$ on the real robot.

\subsection{Ablation Study}

\noindent\textbf{Scene reconstruction and physics solver.}
\Cref{fig:solver_exp} presents qualitative comparisons of scene reconstruction and physics simulation. 
Marker-assisted Reconstruction (\textit{e.g.}, AR Code~\citep{arcode2026argenai}) produces centimeter-level meshes with noticeable artifacts.
In contrast, our pipeline reconstructs assets with sub-millimeter level accuracy, enabling precise geometry alignment for simulation.
Generic deformable solvers (\textit{e.g.}, FEM~\citep{zienkiewicz2005finite}, VBD~\citep{10.1145/3658179}, \textit{etc.}) are not designed for rigid–soft interaction and exhibit unrealistic dynamics due to particle motion lag. This leads to excessive stretching during pulling, particle gaps that cause slipping, and local delays that produce spiky deformations. 
Our solver eliminates these artifacts and yields stable, physically consistent garment behavior, demonstrating the necessity of both \textit{geometric alignment} via accurate scene reconstruction and \textit{dynamical alignment} via stabilized physics simulation.

\noindent \textbf{Quantitative evaluation.}
\Cref{tab:ablation} evaluates module contributions. 
The baseline trajectory strategy, adapted from rigid-body manipulation MimicGen~\citep{mandlekar2023mimicgen}, fails to generate valid training data (pass rate $\textbf{0}\%$), confirming that naive segmentation is insufficient for deformable tasks. 
Trajectory decomposition enables data synthesis but produces discontinuous segments and no task success, indicating limited utility without physical consistency. 
Diffusion-based generation improves data realism (in-domain $\textbf{47}\%$) and captures more human-like trajectories, yet generalization remains weak. 
Incorporating the deformation-stable solver yields substantial gains (in-domain $\textbf{67}\%$, average $\textbf{76}\%$), demonstrating that solver stability is essential for translating scalable simulation data into policies that generalize to the physical world, underscoring the role of \textit{movement alignment} in synthesizing physically consistent manipulation trajectories.

\section{Conclusion}

We present a physics-aligned real-to-sim-to-real pipeline that transforms limited real demonstrations into scalable synthetic data for deformable manipulation. By jointly addressing geometric accuracy, dynamic fidelity, and motion synthesis, our approach achieves zero-shot sim-to-real transfer in garment folding. Experiments demonstrate that policies trained purely on synthetic data achieve comparable performance to real-data baselines, with consistent improvement as data scales. These results show that high-fidelity simulation is a viable and scalable source of supervision for deformable manipulation, complementing real-data collection and enabling broader policy learning at reduced cost.
% Experiments demonstrate that high-quality synthetic data can effectively replace human teleoperation, offering a scalable path toward generalist deformable manipulation.

\noindent \textbf{Limitations and broader impact.} 
A current limitation is that material calibration requires expert-guided parameter tuning for each asset, which constrains full automation across arbitrary cloth types. 
By demonstrating that high-fidelity simulation can complement real-robot data and enable scalable policy learning, this work provides a practical foundation for data-efficient robotic development.

\begin{table}[t]
\centering
\caption{\textbf{Ablation on designs in \modelname{}.} Pass rate denotes the proportion of generated samples accepted by the discriminator. All designs contribute to the final performance.}
\tablestyle{2.5pt}{1.05}
{
\scriptsize
\begin{tabular}{l|c|ccccc|c}
\toprule
\multirow{2}{*}{Method}  & Pass Rate  & \multicolumn{5}{c|}{Success Rate (\%)} & Average \\ 
  & (\%) & In-domain & Spatial & Texture & Lighting & Viewpoint & (\%)\\ \midrule
Baseline &  0  & - & - & - & - & - & - \\ \midrule
+ Traj. decomposition & 65 & 0 & 0 & 0 & 0 & 0 & 0\\ 
+ Diff.-based generation & 38   & 47 & 33 & 60 & 20 & 7& 33 \\ 
\baseline{+ Deform.-stable solver} &  \baseline{40}  & \baseline{\textbf{67}\tiny\textcolor{DarkBlue}{(+20)}} & \baseline{\textbf{93}\tiny\textcolor{DarkBlue}{(+60)}} & \baseline{\textbf{90}\tiny\textcolor{DarkBlue}{(+30)}} & \baseline{\textbf{82}\tiny\textcolor{DarkBlue}{(+62)}} & \baseline{\textbf{47}\tiny\textcolor{DarkBlue}{(+40)}} & \baseline{\textbf{76}\tiny\textcolor{DarkBlue}{(+43)}}\\ \bottomrule
\end{tabular}
}
\label{tab:ablation}
\end{table}

\section*{Ackonwledgement}
We sincerely thank Jiafei Cao, Yang Li, and Junjie Xia for their invaluable support in building and maintaining the data pipeline. We are also grateful to all data collection contributors for their dedicated efforts in large-scale data acquisition.
We appreciate Chaoyang Lv for his assistance with the simulation system and related technical support.
We thank Haochen Tian for developing the visualization and plotting scripts used in this work. 
We also acknowledge the support and guidance from Prof. Minyi Guo and Prof. Hongzi Zhu.

\bibliography{refs}

\appendix
\newpage

\section{Author contributions}
\label{sec:authors}
All authors contributed to writing.

Contributors: Yunsong Zhou (zhouyunsong@pjlab.org.cn), Hangxu Liu, Xuekun Jiang, Xing Shen, Yuanzhen Zhou, Hui Wang, Baole Fang, and Yang Tian.

Advisors: Jiangmiao Pang (pangjiangmiao@gmail.com), Mulin Yu, Qiaojun Yu, Li Ma, Hengjie Li, Hanqing Wang, and Jia Zeng.

% \begin{itemize}[leftmargin=*]
%     \item Yunsong Zhou (zhouyunsong@pjlab.org.cn): Proposed and led project.
%     \item Hangxu Liu: Contributed to simulation infrastructure, simulated data generation, as well as open-sourcing.
%     \item Xuekun Jiang: Led simulation infrastructure and open-sourcing. Contributed to the validity checker and experiments.
%     \item Xing Shen: Led simulation solver development.
%     \item Yuanzhen Zhou: Contributed to scene randomization and rendering.
%     \item Hui Wang: Contributed to early versions of simulators.
%     \item Baole Fang: Contributed to scene digitalization.
%     \item Yang Tian: Provided suggestions and contributed to outreach efforts.
%     \item Mulin Yu: Advised asset scanning.
%     \item Qiaojun Yu: Advised project. Contributed to experiments. 
%     \item Li Ma: Advised infrastructures.
%     \item Hengjie Li: Advised infrastructures.
%     \item Hanqing Wang: Advised project. Supported computational resources. 
%     \item Jia Zeng: Advised project. Contributed to data collection pipelines.
%     \item Jiangmiao Pang (pangjiangmiao@gmail.com): Supervised project direction with critical feedback.
% \end{itemize}

\section{\modelletter{} Implementation Details}
\label{sec:avbd_implementation}

In our framework, we use the Augmented Vertex Block Descent (AVBD)~\citep{10.1145/3731195} solver to simulate cloth dynamics. 

\subsection{Solver Overview}

Our cloth solver operates on a triangular mesh where edges $\mathbf{e}=(\mathbf{e}_i, \mathbf{e}_j)$ represent connections between vertices. 
The goal of each Newton-type iteration is to compute a displacement update $\Delta \mathbf{e}_i$ for each vertex such that the mesh configuration minimizes the total energy while satisfying constraints (stretch limits, bending, collision avoidance).
The overall workflow is summarized as follows:

\begin{enumerate}
    \item Compute internal forces derived from elastic and bending energies.
    \item Evaluate constraint violations (stretch and strain) and compute corrective forces \textbf{(new)}.
    \item Assemble total forces acting on each vertex.
    \item Solve for vertex updates $\Delta \mathbf{e}_i$ using the Newton system.
    \item Apply penetration avoidance to truncate updates that would violate collision safety \textbf{(new)}.
    \item Iterate until convergence, producing the new vertex positions $\mathbf{e}_i^{\text{new}}$.
\end{enumerate}

\subsection{Internal Force Computation}

\noindent \textbf{StVK elasticity model.} 
To simulate how cloth resists stretching and shearing, we employ the \textit{St. Venant--Kirchhoff (StVK)} hyperelastic model~\citep{10.1145/1477926.1477934}. Intuitively, this model assigns an energy to how much each triangle in the cloth mesh is deformed compared to its rest shape: the more a triangle stretches or shears, the higher the energy. Minimizing this energy naturally generates forces that restore the cloth toward its undeformed configuration.

Formally, let $\mathbf{e}$ denote the positions of the mesh vertices, and let $\mathbf{F}$ be the deformation gradient mapping rest-state coordinates to the current positions. The Green-Lagrange strain tensor
$\mathbf{G} = \frac{1}{2}(\mathbf{F}^T \mathbf{F} - \mathbf{I})$
measures the nonlinear strain, where $\mathbf{I}$ is the identity matrix. The elastic energy density of a triangle is then:
\begin{equation}
\Psi(\mathbf{e}) = \mu \|\mathbf{G}\|_F^2 + \frac{1}{2} \hat{\lambda}\, (\text{tr}(\mathbf{G}))^2,
\end{equation}
where $\|\cdot\|_F$ denotes the Frobenius norm, $\text{tr}(\cdot)$ is the trace operator, and $\mu$ and $\hat{\lambda}$ are the Lam\'e parameters that control the material's resistance to shear and area change, respectively.

The elastic (stretching) force acting on vertex $i$ is obtained as the negative gradient of the energy with respect to its position $\mathbf{f}_i^{\text{stretch}} = - \frac{\partial \Psi}{\partial \mathbf{e}_i}$.
Intuitively, this force acts like a spring that resists local stretching and shearing, driving the cloth mesh toward physically plausible configurations.

\noindent \textbf{Dihedral-angle bending model.}  
To capture out-of-plane deformations such as folding or wrinkling, we adopt a \textit{dihedral-angle} bending model~\citep{zhou2008plausible}. Conceptually, each edge shared by two adjacent triangles has a preferred rest angle; deviations from this angle induce a restoring force that resists bending.

Let $\mathbf{e}_i, \mathbf{e}_j, \mathbf{e}_k, \mathbf{e}_l$ denote the four vertices forming the two triangles sharing an edge. The current dihedral angle $\theta$ is a function of these vertex positions, and $\theta_0$ is the rest angle. Denoting the edge length by $l_{\text{edge}}$ and bending stiffness by $k_{\text{bend}}$, the bending energy is:
\begin{equation}
E_{\text{bend}}(\mathbf{e}_i, \mathbf{e}_j, \mathbf{e}_k, \mathbf{e}_l) = 
k_{\text{bend}} \, l_{\text{edge}} \, (\theta(\mathbf{e}_i, \mathbf{e}_j, \mathbf{e}_k, \mathbf{e}_l) - \theta_0)^2.
\end{equation}

The bending force on vertex $\mathbf{e}_i$ is $\mathbf{f}_i^{\text{bend}} = - \frac{\partial E_{\text{bend}}}{\partial \mathbf{e}_i}$, and similarly for the other vertices of the edge. Each vertex accumulates contributions from all adjacent edges, which combined with the stretching and constraint forces to form the total force used by the solver to compute the vertex displacement update $\Delta \mathbf{e}_i$.

\subsection{Constraint Force Computation}

\noindent \textbf{Spring strain constraint.}  
To suppress unphysical stretching, we enforce a maximum stretch ratio $\xi$ on all mesh edges, as formulated in \Cref{sec:real_to_sim_matching} (\Cref{eq:constraint}-\Cref{eq:energy}) in the main paper.
The resulting force on vertex \(\mathbf{e}_i\) is computed as $\mathbf{f}_i^{\text{strain}} = - \frac{\partial E_{\text{strain}}}{\partial \mathbf{e}_i}$, and similarly for \(\mathbf{e}_j\). These constraint forces are then integrated with stretching, bending, and external forces to determine the total vertex displacement \(\Delta \mathbf{e}_i\) during each AVBD iteration.

\subsection{Total Force Assembly}

The vertex update \(\Delta \mathbf{e}_i\) is computed by solving a local \(3\times 3\) linear system derived from the Newton optimization~\citep{10.1145/3197517.3201308}. Specifically, we define the system matrix \(\mathbf{A}_i\) and the residual vector \(\mathbf{b}_i\) as:
\begin{equation}
\mathbf{A}_i = \frac{M_i}{\Delta t^2} \mathbf{I} + \sum \mathbf{H}_i, \quad
\mathbf{b}_i = \mathbf{f}_i^{\text{total}} - \frac{M_i}{\Delta t^2}(\mathbf{e}_i^{(n)} - \hat{\mathbf{e}}_i),
\end{equation}
where \(M_i\) is the lumped mass of vertex \(i\), \(\Delta t\) is the time step, and \(\mathbf{H}_i\) is the sum of Hessians of the energy terms associated with vertex \(i\)~\citep{10.1145/2366145.2366171}, $\mathbf{e}_i^{(n)}$ is the vertex position at the current Newton-type iteration $n$, $\hat{\mathbf{e}}_i$ is the inertial predictive position, a constant reference for the current time step derived from the vertex's velocity and acceleration in the preceding time frame.

The total force \(\mathbf{f}_i^{\text{total}}\) is assembled from multiple contributions:
\begin{equation}
\mathbf{f}_i^{\text{total}} = 
\mathbf{f}_i^{\text{stretch}} + 
\mathbf{f}_i^{\text{bend}} + 
\mathbf{f}_i^{\text{strain}}  + 
\mathbf{f}_i^{\text{ext}},
\end{equation}
where \(\mathbf{f}_i^{\text{stretch}}\) comes from StVK elasticity, \(\mathbf{f}_i^{\text{bend}}\) from dihedral-angle bending, \(\mathbf{f}_i^{\text{strain}}\) from the spring strain constraint, and \(\mathbf{f}_i^{\text{ext}}\) includes any external forces applied by robots or environment interactions.
The displacement is obtained by solving the linear system, $\Delta \mathbf{e}_i = \mathbf{A}_i^{-1} \mathbf{b}_i$.

\subsection{Penetration Avoidance} 
To prevent interpenetration with obstacles or self-collision, we introduce a geometric safety filter. 
A safe displacement bound $d_{\text{safe}}$ is defined as $d_{\text{safe}} = \beta \cdot \min(r_{\text{collision}}, d_{\text{tri}}^{\min}, d_{\text{edge}}^{\min})$~\citep{10.1145/566654.566623}, where $\beta=0.42$ is a relaxation factor, $r_{\text{collision}}$ is the collision radius, and $d_{\text{tri}}^{\min}, d_{\text{edge}}^{\min}$ represent the minimum distances to the nearest triangle and edge primitives, respectively. Based on this bound, a scalar clipping factor $s$ is computed to truncate the raw Newton update $\Delta \mathbf{e}_i$:
\begin{equation}
s = \min\left(1, \frac{d_{\text{safe}}}{\|\Delta \mathbf{e}_i\|}\right).
\end{equation}
This mechanism ensures that the vertex does not overshoot the safety margin during a single iteration, effectively preventing tunneling and numerical instability.

\subsection{Position Updation}

The sequence concludes with the synthesis of the final vertex position $\mathbf{e}_i^{\text{new}}$ for the current iteration. By applying the clipped displacement to the iteration's starting point $\mathbf{e}_i^{(n)}$, we obtain:
\begin{equation}
\mathbf{e}_i^{\text{new}} = \mathbf{e}_i^{(n)} + s \cdot \Delta \mathbf{e}_i.
\end{equation}
This integration ensures that the updated mesh state is not only physically optimal according to the AVBD energy gradients but also strictly compliant with geometric safety constraints, leading to a robust and collision-free simulation.

\begin{table}[t]
\centering
\caption{\textbf{Calibrated simulation parameters for cloth physics in \modelname{}}.}
\label{tab:cloth_parameters}
\footnotesize
\renewcommand{\arraystretch}{1.2} 
\tablestyle{20.5pt}{1.05}
\begin{tabular}{l|c|c}
\toprule
\textbf{Category} & \textbf{Parameter} & \textbf{Value} \\
\midrule
\multirow{2}{*}{Particle} & Radius & 0.008 m \\
                          & Density & 2.0 kg/m$^2$ \\
\midrule
\multirow{3}{*}{StVK Elasticity} & Shear modulus $\mu$ & $1.0 \times 10^2$ \\
                                  & Area modulus $\lambda$ & $1.0 \times 10^2$ \\
                                  & Damping & $1.5 \times 10^{-6}$ \\
\midrule
\multirow{2}{*}{Bending} & Stiffness & $8.0 \times 10^{-4}$ \\
                         & Damping & $1.0 \times 10^{-3}$ \\
\midrule
\multirow{4}{*}{Strain Limit} & Maximum stretch ratio & 0.05 (5\%) \\
                              & Initial stiffness & $1.0 \times 10^4$ \\
                              & Maximum stiffness & $1.0 \times 10^6$ \\
                              & Growth rate & $1.0 \times 10^3$ \\
\midrule
\multirow{8}{*}{Contact} & Soft contact stiffness & $5.0 \times 10^2$ \\
                         & Soft contact damping & $5.0 \times 10^{-3}$ \\
                         & Robot friction & 1.5 \\
                         & Table friction & 0.0 \\
                         & Self-contact friction & 0.25 \\
                         & Self-contact radius & 0.002 m \\
                         & Body-cloth margin & 0.01 m \\
                         & Friction smoothing & $1.0 \times 10^{-2}$ \\
\midrule
\multirow{4}{*}{AVBD Temporal} & Lambda decay & 0.94 \\
                               & Stiffness decay & 0.95 \\
                               & Regularization $\alpha$ & 0.99 \\
                               & Bound relaxation $\beta$ & 0.42 \\
\bottomrule
\end{tabular}
\end{table}

\subsection{Simulation Infrastructure}
\label{sec:simulation_infrastructure}

To support high-fidelity data collection for deformable manipulation, we develop a simulation infrastructure. This pipeline enables real-time synchronization between physical teleoperation and GPU-accelerated simulation, featuring a tight integration of rigid-body and cloth dynamics.

\noindent \textbf{Bidirectional teleoperation and joint mapping.}
During teleoperation, we establish a direct correspondence between the physical robot and the simulated actuators. Specifically, the simulation joint state $\mathbf{q}_{\text{sim}}$ is updated as:
\begin{equation}
\mathbf{q}_{\text{sim}}(t)[\mathcal{I}_{\text{sim}}] = \mathbf{q}_{\text{real}}(t)[\mathcal{I}_{\text{arm}}],
\label{eq:joint_mapping}
\end{equation}
where $\mathcal{I}_{\text{sim}}$ and $\mathcal{I}_{\text{arm}}$ represent the index sets of controllable joints in the simulator and the physical robot arm, respectively. For the end-effectors, the left and right gripper openness ($o_L, o_R \in [0, 1]$) are decoupled and mapped independently from their respective physical finger joints to ensure asymmetric grasping fidelity:
\begin{equation}
o_L = -\text{clip}\left(\frac{q_{\text{finger}, L}}{2 \cdot q_{\text{max}}}, -1, 0\right), \quad o_R = -\text{clip}\left(\frac{q_{\text{finger}, R}}{2 \cdot q_{\text{max}}}, -1, 0\right),
\label{eq:gripper_normalization}
\end{equation}
where $q_{\text{finger}, L}$ and $q_{\text{finger}, R}$ are the joint angles of the left and right physical fingers, and $q_{\text{max}} = 1.62$ rad is the mechanical limit.

\noindent \textbf{Physical parameter calibration.} 
We calibrate the cloth's physical properties within the AVBD framework to ensure realism and stability, as summarized in Table~\ref{tab:cloth_parameters}. This includes StVK moduli for elasticity, a 5\% strain limit to prevent over-stretching, and contact parameters such as robot-specific friction and the bound relaxation factor $\beta=0.42$. These parameters are specifically tuned to maintain numerical robustness during the high-speed, contact-rich interactions characteristic of \modelname{} tasks.

\noindent \textbf{GPU-accelerated simulation.}
The simulation is powered by NVIDIA Warp, a high-performance framework that compiles Python code into native CUDA kernels for GPU execution. This allows our rigid-body dynamics and AVBD cloth solver to run entirely on the GPU within a unified memory space ($\sim$15 fps).

\noindent \textbf{Data recording.}
The system employs asynchronous logging to maintain simulation throughput. Per-frame robot states and gripper openness are stored in NPZ format for policy learning, while full session trajectories and contact manifolds are serialized to USD for post-hoc diagnostic analysis and visual inspection.

\begin{algorithm}[t]
\caption{Synthetic data generation via trajectory decomposition and diffusion-based motion generation}
\label{alg:sim_scene_generation_supp}
\begin{algorithmic}[1]
\Require Expert demonstrations $\mathcal{D}$, simulator $\mathcal{E}$, diffusion model $f_\theta$, discriminator $D$, renderer $\mathcal{R}$, appearance count $K$
\Ensure Synthetic dataset $\mathcal{D}_{\text{synth}}$
\State Extract interaction segments $\mathcal{P} = \{(\mathbf{p}_s^i, \mathbf{p}_t^i)\}_{i=1}^N$ from $\mathcal{D}$ \Comment{Structured decomposition}
\While{$|\mathcal{D}_{\text{synth}}| < N_{\text{target}}$}
    \State Sample segment sequence $\{p_1,\dots,p_L\} \sim \mathcal{P}$ \Comment{Decomposed task skeleton}
    \State $\tau_{\text{gen}} \gets \emptyset$
    \For{$k = 1$ to $L-1$}
        \State $\mathbf{p}_s \gets \text{end}(p_k)$, $\mathbf{p}_t \gets \text{start}(p_{k+1})$
        \State $\mathbf{h} \sim \mathcal{D}$ \Comment{Sample demonstration history}
        \State $\mathbf{x}_{1:M} \gets f_\theta(\mathbf{h}, \mathbf{p}_s, \mathbf{p}_t)$ \Comment{Diffusion-based trajectory synthesis}
        \State $\tau_{\text{gen}} \gets \tau_{\text{gen}} \cup \mathbf{x}_{1:M}$
    \EndFor
    \State Execute $\tau_{\text{gen}}$ in $\mathcal{E}$, obtain video $\mathbf{V}$
    \If{$\text{success}(\tau_{\text{gen}})$ and $D(\mathbf{V}) > \tau$}
        \For{$v = 1$ to $K$}
            \State $\mathbf{V}_v \gets \mathcal{R}(\tau_{\text{gen}})$ with randomized appearance
            \State Store $(\tau_{\text{gen}}, \mathbf{V}_v)$ in $\mathcal{D}_{\text{synth}}$
        \EndFor
    \EndIf
\EndWhile
\State \Return $\mathcal{D}_{\text{synth}}$
\end{algorithmic}
\end{algorithm}

\section{Synthetic Data Generation Algorithm}
\label{sec:appendix_sim_data_gen}

\Cref{alg:sim_scene_generation_supp} presents the pseudo-code of our synthetic data generation pipeline, which transforms teleoperated demonstrations into large-scale synthetic trajectories through structured decomposition and diffusion-based trajectory synthesis.

The process begins by extracting reusable interaction segments from expert demonstrations (line 1), forming a library of grasp-to-release primitives that capture meaningful manipulation phases. During data generation, sequences of these segments are sampled to construct a high-level task skeleton (lines 2–3).
For each adjacent segment pair, a diffusion model synthesizes feasible transition trajectories that connect the segment endpoints (lines 5–9), producing a complete manipulation trajectory. The generated trajectory is then executed in the simulator to obtain a corresponding video sequence (line 10).
To ensure data quality, trajectories are filtered using both task success checks and a discriminator that evaluates visual realism (line 11). Finally, valid trajectories are rendered multiple times with randomized appearances (lines 12–14), producing diverse observations while sharing the same underlying physical interaction.
This procedure is repeated until the desired dataset size is reached.

\begin{table}[t]
\centering
\footnotesize
\setlength{\tabcolsep}{10pt}
\caption{\textbf{Generalization performance on additional tasks.} $\dagger$: Out-of-domain experiments where policies are evaluated on entirely unseen real-world tasks.}
\label{tab:generalization}
\renewcommand{\arraystretch}{1.2}
\begin{tabular}{l|cc|c}
\toprule
Task & Towel Flipping & Shorts Folding & Polo-shirt Folding$^{\dagger}$ \\
\midrule
Success rate (\%) & \textbf{80} & \textbf{93} & \textbf{93} \\
\bottomrule
\end{tabular}
\end{table}

\section{Experiments}

\subsection{Additional Tasks}

To demonstrate generalizability beyond T-shirts, we apply our pipeline to towels and shorts. For each category, interaction primitives are manually designed. Starting with 100 teleoperated demonstrations per category, we generate 1,000 synthetic trajectories with texture randomization.
We evaluate policies trained on the synthetic datasets over 30 real-world trials per task, achieving 80\% success on towel folding and 93\% on shorts and polo-shirt folding (\Cref{tab:generalization}). While towels and shorts require teleoperated demonstrations for data generation, the polo-shirt result is achieved in a \textbf{zero-shot} manner without any task-specific demonstrations.

Notably, the polo-shirt exhibits substantially different geometry, size, material, and frictional properties compared to the garments used during training. Moreover, no similar garment instance appears in the training dataset. Despite this significant distribution shift, the learned policy transfers successfully to the real-world task, highlighting strong cross-garment generalization.

\begin{figure}[t]
    \centering
    \includegraphics[width=0.99\linewidth]{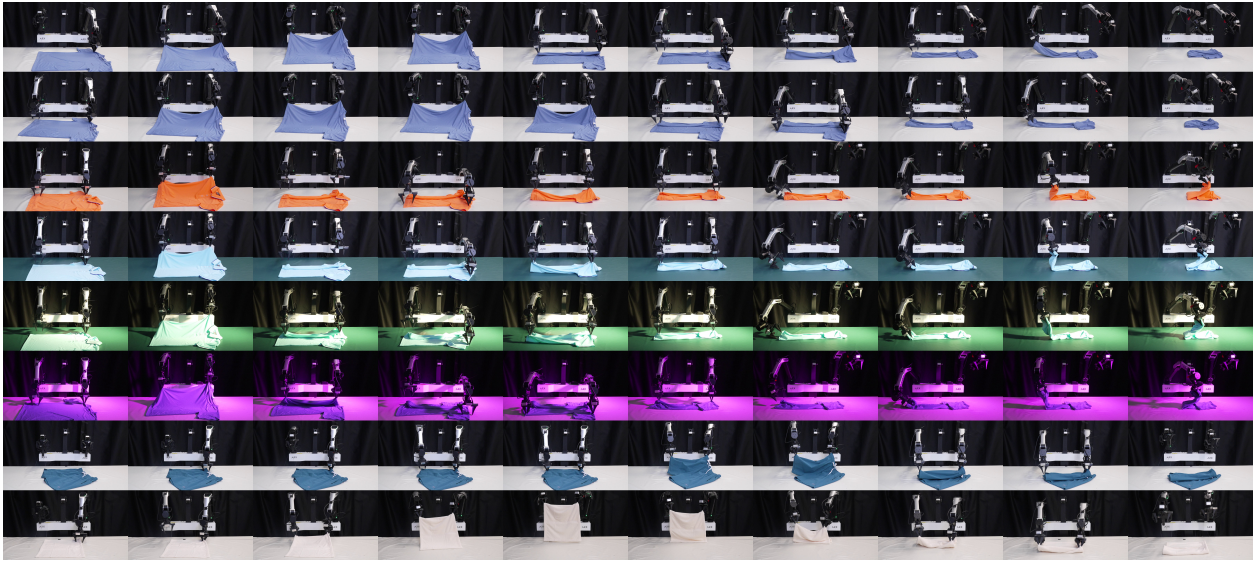}
    \caption{\textbf{Real-world deployment experiments.} Representative results include T-shirt folding under both in-domain and out-of-domain settings. We further demonstrate generalization to additional garments, including polo-shirts, shorts, and towels.}
    \label{fig:video_grid_real}
\end{figure}

\subsection{Real-world Deployment}

\Cref{fig:video_grid_real} shows representative real-world deployment results. Each row corresponds to a different experiment. From top to bottom, we present in-domain T-shirt folding, T-shirt folding with randomized object positions, polo-shirt folding under varying textures and lighting conditions, and experiments on shorts and towels. The viewpoint randomization setting is not shown, as it produces minimal visual differences.

\begin{figure}[htbp]
    \centering
    \includegraphics[width=0.99\linewidth]{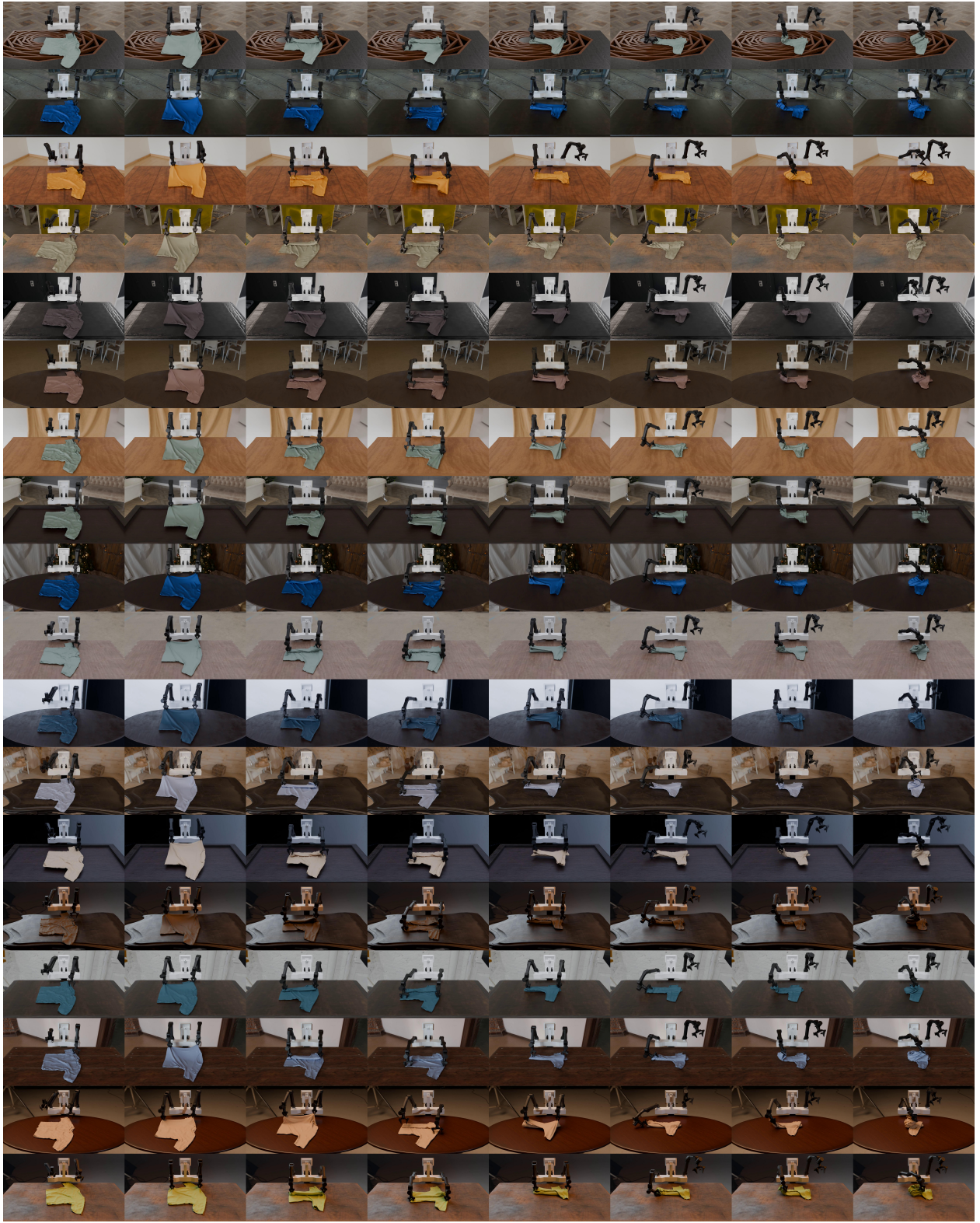}
    \caption{\textbf{Simulated T-shirt folding scenarios.} Each row shows a distinct configuration generated by our pipeline.}
    \label{fig:supp_data}
\end{figure}

\begin{figure}[htbp]
    \centering
    \includegraphics[width=0.99\linewidth]{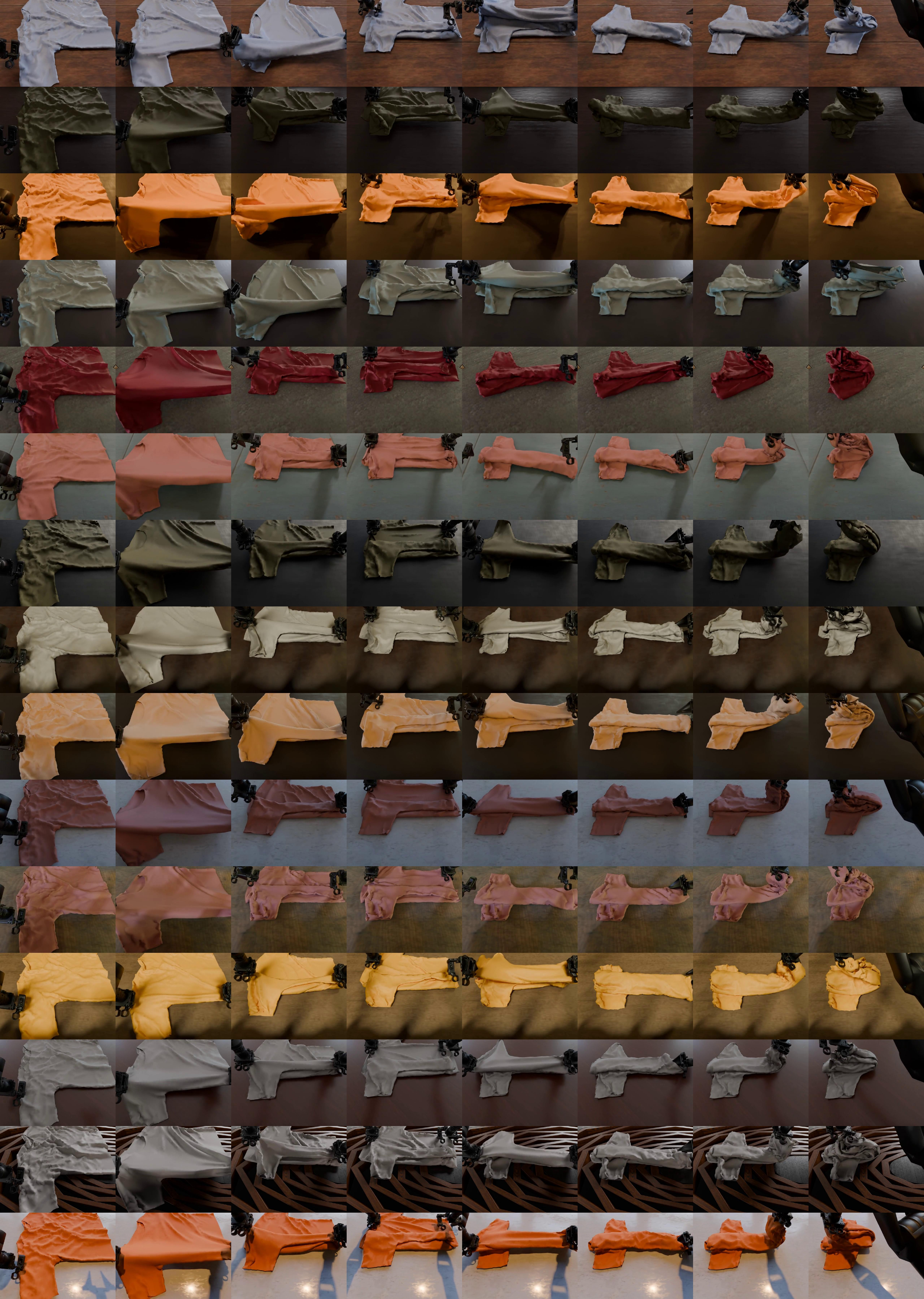}
    \caption{\textbf{Synthetic generated data of T-shirt folding.} Each row shows a temporally sampled trajectory with randomized textures, illustrating the visual diversity generated by our pipeline.}
    \label{fig:video_grid}
\end{figure}

\subsection{More Visualizations}

\begin{figure}[htbp]
    \centering
    \includegraphics[width=0.99\linewidth]{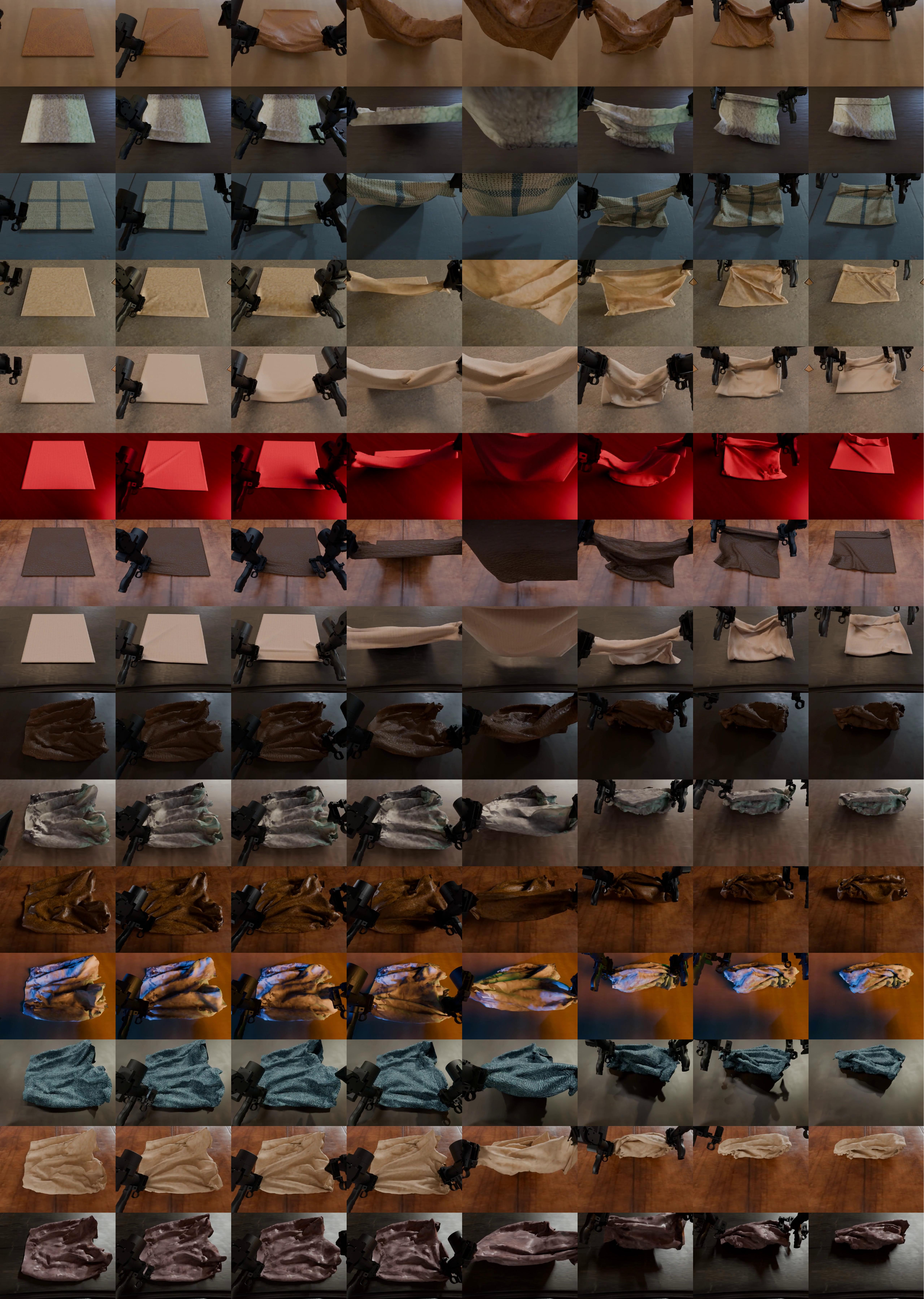}
    \caption{\textbf{Synthetic generated data of towels and shorts.} Each row shows a temporally sampled trajectory with randomized textures, demonstrating the framework's applicability to multiple garment types.}
    \label{fig:video_grid_merged}
\end{figure}

We visualize synthetic demonstrations generated by our pipeline to highlight both diversity and realism. \Cref{fig:supp_data} shows representative T-shirt folding scenes with randomized garments, tables, and lighting conditions. 
Beyond T-shirts, we also illustrate towel, shorts, and polo-shirt folding to demonstrate generality. \Cref{fig:video_grid} and \Cref{fig:video_grid_merged} present temporally sampled sequences captured from head-mounted cameras, aligned with real-robot data, showcasing the diversity achieved through appearance and scene randomization.

\subsection{Cost-efficiency Analysis}
\label{sec:cost-efficiency-analysis}

We conduct a comparative analysis to quantify the economic and throughput advantages of our simulation pipeline against traditional real-world data acquisition in \Cref{tab:cost_analysis}. 
Real-world data collection incurs a daily cost of approximately \$282, comprising \$200 for manual labor (8 hours at \$25/h) and \$82 for hardware depreciation (calculated for a \$30,000 platform over a one-year lifecycle). 
Given an average yield of 104 trajectories per day, the unit cost amounts to approximately \$2.71 per trajectory.

In contrast, our simulation framework, deployed on a server equipped with $8 \times$ NVIDIA RTX 4090 GPUs, operates at a daily cost of roughly \$71 (amortized to \$0.37 per hour). 
With an average rendering time of 16.2 minutes per GPU, the system enables the parallel generation of approximately 710 trajectories daily, reducing the unit cost to \$0.10 per trajectory. 
This represents a $27\times$ reduction in cost and a $6.8\times$ increase in throughput compared to physical data collection. 
These results demonstrate that physics-aligned simulation provides a highly scalable and cost-efficient alternative for generating large-scale training data.

\begin{table}[t]
\caption{\textbf{Comparison of data acquisition efficiency.} All costs are estimated based on the daily expense of employing two data operators (\$282).}
\label{tab:cost_analysis}
\centering
\footnotesize
\begin{tabular}{l|ccc}
\toprule
Method & Throughput (traj./day)$\uparrow$ & Unit Cost (\$/traj.)$\downarrow$ & Relative Cost $\downarrow$\\ \midrule
Real-world Collection & 104 & 2.71 & $1.0\times$ \\
Ours (Simulation) & \textbf{710} & \textbf{0.10} & $\mathbf{0.037\times}$ \\ \bottomrule
\end{tabular}
\end{table}

\subsection{Failure Cases}

Our experiments indicate that policy performance is highly sensitive to the quality of generated data. When low-quality or invalid samples enter the training set, due to discriminator errors, they can poison the model, leading to systematic failures. Specifically, corrupted samples often result in unreasonable behaviors such as overreaching, premature gripper closure before contact, or misaligned grasps. In contrast, genuine out-of-distribution scenarios primarily produce failures due to limited generalization, such as missing the garment edge or mispositioning relative to the object. In our pipeline, the discriminator achieves over 99\% success in filtering invalid trajectories, and its performance can be further improved with simple rule-based checks, ensuring that the synthetic dataset remains highly reliable and minimally corrupted.

\end{document}